\documentclass[10pt,twocolumn,letterpaper]{article}

\usepackage{cvpr}
\usepackage{times}
\usepackage{epsfig}
\usepackage{graphicx}
\usepackage{amsmath}
\usepackage{amssymb}
\usepackage{float}

\newcommand{\finalcopy}{\cvprfinalcopy}
\usepackage{tikz}
\usetikzlibrary{arrows,shapes,calc,matrix,fit,positioning,backgrounds}
\usepackage{pgfplots}
\usepackage{pgfplotstable}
\pgfplotsset{compat=1.9}

\usepackage{xstring}

\usepgfplotslibrary{external}
\newcommand{\extfig}[2]{\tikzsetnextfilename{fig/extern/#1}{#2}}

\IfBeginWith*{\jobname}{fig/extern/}{\finalcopy}{}

\newcommand{\leg}[1]{\addlegendentry{#1}}

\tikzset{every mark/.append style={solid}}
\pgfplotsset{
	grid=both, width=\columnwidth, try min ticks=5,
	every axis/.append style={font=\scriptsize},
	every axis plot/.append style={thick,mark=none,mark size=1.2,tension=0.18},
	legend cell align=left, legend style={fill opacity=0.8},
}

\pgfplotsset{
	dash/.style={mark=o,dashed,opacity=0.7},
	dott/.style={mark=o,dotted,opacity=0.7},
}

\newcommand{\map}[1]{\thisrow{#1}*100}

\tikzstyle{every picture}+=[
	remember picture,
	every text node part/.style={align=center},
	every matrix/.append style={ampersand replacement=\&},
]
\tikzstyle{tight} = [inner sep=0pt,outer sep=0pt]
\tikzstyle{txt} = [inner sep=1pt,outer sep=0pt,anchor=base]
\tikzstyle{base} = [baseline,every node/.style={tight,anchor=base}]
\tikzstyle{lab} = [draw=none,fill=blue!10,anchor=base,font=\footnotesize]
\tikzstyle{label} = [baseline,every node/.style={lab}]
\tikzstyle{over} = [overlay,every path/.style={->,draw=red,thick},nodes={draw}]
\tikzstyle{semi} = [draw opacity=.5]
\tikzstyle{paper} = [fill=white]
\tikzstyle{help} = [gray]
\tikzstyle{high} = [blue!70]
\tikzstyle{op} = [draw,circle,tight]
\tikzstyle{dot} = [fill,draw,circle,inner sep=1pt,outer sep=0]
\tikzstyle{circ} = [draw,circle,thick,inner sep=1.5pt,outer sep=0]
\tikzstyle{ocirc} = [draw,circle,thick,inner sep=2pt,outer sep=0]
\tikzstyle{node} = [draw,circle,tight,minimum size=12pt,anchor=center]
\tikzstyle{box} = [draw,thick,rectangle,inner sep=6pt]
\tikzstyle{group} = [high,box,opacity=.5]
\tikzstyle{maths} = [matrix of math nodes]
\tikzstyle{rsep} = [row sep={#1,between origins}]
\tikzstyle{csep} = [column sep={#1,between origins}]
\tikzstyle{short} = [shorten >=#1,shorten <=#1]
\tikzstyle{dim1} = [fill opacity=.3,text opacity=1]
\tikzstyle{dim2} = [fill opacity=.5,text opacity=1]
\tikzstyle{dim3} = [fill opacity=.7,text opacity=1]
\tikzstyle{rectc} = [tight,transform shape]
\tikzstyle{rect} = [rectc,anchor=south west]

\def\bx[#1,#2]{#1}
\def\by[#1,#2]{#2}
\def\dx[#1,#2]{#1,0}
\def\dy[#1,#2]{0,#2}

\pgfkeys{/tikz/.cd, aspect/.store in=\aspect, aspect=3}
\pgfkeys{/tikz/.cd, depth/.store in=\depth, depth=.15}
\pgfkeys{/tikz/.cd, stepx/.store in=\step, stepx=1}

\tikzstyle{geom} = [line join=bevel,aspect=3,depth=.25,z={(\depth*\aspect,\depth)}]
\tikzstyle{wire} = [geom,draw,thick]

\def\cx[#1,#2,#3]{#1}
\def\cy[#1,#2,#3]{#2}
\def\cz[#1,#2,#3]{#3}
\def\ex[#1,#2,#3]{#1,0,0}
\def\ey[#1,#2,#3]{0,#2,0}
\def\ez[#1,#2,#3]{0,0,#3}

\newcommand{\xell}[4][]{%
\path[geom,#1,shift={#2},yslant=1/\aspect]
	(0,0) ellipse [x radius={\bx[#3]cm},y radius={\by[#3]cm},rotate=#4];
}

\newcommand{\zrect}[3][]{%
\path[geom,#1] #2 rectangle +(\cx[#3],\cy[#3]);
}
\newcommand{\yrect}[3][]{%
\path[geom,#1,shift={#2},xslant=\aspect]
	(0,0) rectangle +(\cx[#3],\depth*\cz[#3]);
}
\newcommand{\xrect}[3][]{%
\path[geom,#1,shift={#2},yslant=1/\aspect]
	(0,0) rectangle +(\aspect*\depth*\cz[#3],\cy[#3]);
}

\newcommand{\para}[3][]{%
\zrect[#1]{(#2)}{#3}                 
\yrect[#1]{($(#2)+(\ey[#3])$)}{#3}   
\xrect[#1]{($(#2)+(\ex[#3])$)}{#3}   
}

\newcommand{\xrectim}[4][]{%
\path[geom,#1,shift={#2},yslant=1/\aspect]
	let \n0={\aspect*\depth*\cz[#3]} in
		(0,0) node[rect] {\includegraphics[width={\n0cm},height={\cy[#3]cm}]{#4}};
}
\newcommand{\paraim}[4][]{%
\zrect[#1]{(#2)}{#3}                       
\yrect[#1]{($(#2)+(\ey[#3])$)}{#3}         
\xrectim[#1]{($(#2)+(\ex[#3])$)}{#3}{#4}   
\xrect[#1]{($(#2)+(\ex[#3])$)}{#3}         
}

\usepackage{enumitem}
\usepackage{multirow}

\usepackage[pagebackref=true,breaklinks=true,letterpaper=true,colorlinks,bookmarks=false]{hyperref}

\cvprfinalcopy 

\def\cvprPaperID{5468} 

\ifcvprfinal\fi
\begin{document}

\title{Local Features and Visual Words Emerge in Activations}

\author{
Oriane Sim\'eoni$^1$ \ \ \ \ Yannis Avrithis$^1$\ \ \ \ Ond{\v r}ej Chum$^2$\\
{\fontsize{11}{13}\selectfont$^1$Univ Rennes, Inria, CNRS, IRISA \ \ \ \ \ \ $^2$VRG, FEE, Czech Technical University in Prague}
}

\maketitle

\newcommand{\head}[1]{{\smallskip\noindent\bf #1}}
\newcommand{\alert}[1]{{\color{black}{#1}}}
\newcommand{\eq}[1]{(\ref{eq:#1})\xspace}

\newcommand{\red}[1]{{\color{red}{#1}}}
\newcommand{\blue}[1]{{\color{blue}{#1}}}
\newcommand{\green}[1]{{\color{green}{#1}}}
\newcommand{\gray}[1]{{\color{gray}{#1}}}


\newcommand{\tran}{^\top}
\newcommand{\mtran}{^{-\top}}
\newcommand{\zcol}{\mathbf{0}}
\newcommand{\zrow}{\zcol\tran}

\newcommand{\ind}{\mathbbm{1}}
\newcommand{\expect}{\mathbb{E}}
\newcommand{\nat}{\mathbb{N}}
\newcommand{\zahl}{\mathbb{Z}}
\newcommand{\real}{\mathbb{R}}
\newcommand{\proj}{\mathbb{P}}
\newcommand{\prob}{\mathbf{Pr}}

\newcommand{\mif}{\textrm{if }}
\newcommand{\minimize}{\textrm{minimize }}
\newcommand{\maximize}{\textrm{maximize }}
\newcommand{\st}{\textrm{subject to }}

\newcommand{\id}{\operatorname{id}}
\newcommand{\const}{\operatorname{const}}
\newcommand{\sgn}{\operatorname{sgn}}
\newcommand{\var}{\operatorname{Var}}
\newcommand{\mean}{\operatorname{mean}}
\newcommand{\trace}{\operatorname{tr}}
\newcommand{\diag}{\operatorname{diag}}
\newcommand{\vect}{\operatorname{vec}}
\newcommand{\cov}{\operatorname{cov}}

\newcommand{\softmax}{\operatorname{softmax}}
\newcommand{\clip}{\operatorname{clip}}

\newcommand{\defn}{\mathrel{:=}}
\newcommand{\peq}{\mathrel{+\!=}}
\newcommand{\meq}{\mathrel{-\!=}}

\newcommand{\floor}[1]{\left\lfloor{#1}\right\rfloor}
\newcommand{\ceil}[1]{\left\lceil{#1}\right\rceil}
\newcommand{\inner}[1]{\left\langle{#1}\right\rangle}
\newcommand{\norm}[1]{\left\|{#1}\right\|}
\newcommand{\frob}[1]{\norm{#1}_F}
\newcommand{\card}[1]{\left|{#1}\right|\xspace}
\newcommand{\diff}{\mathrm{d}}
\newcommand{\der}[3][]{\frac{d^{#1}#2}{d#3^{#1}}}
\newcommand{\pder}[3][]{\frac{\partial^{#1}{#2}}{\partial{#3^{#1}}}}
\newcommand{\ipder}[3][]{\partial^{#1}{#2}/\partial{#3^{#1}}}
\newcommand{\dder}[3]{\frac{\partial^2{#1}}{\partial{#2}\partial{#3}}}

\newcommand{\wb}[1]{\overline{#1}}
\newcommand{\wt}[1]{\widetilde{#1}}

\def\xssp{\hspace{1pt}}
\def\ssp{\hspace{3pt}}
\def\msp{\hspace{5pt}}
\def\lsp{\hspace{12pt}}

\newcommand{\cA}{\mathcal{A}}
\newcommand{\cB}{\mathcal{B}}
\newcommand{\cC}{\mathcal{C}}
\newcommand{\cD}{\mathcal{D}}
\newcommand{\cE}{\mathcal{E}}
\newcommand{\cF}{\mathcal{F}}
\newcommand{\cG}{\mathcal{G}}
\newcommand{\cH}{\mathcal{H}}
\newcommand{\cI}{\mathcal{I}}
\newcommand{\cJ}{\mathcal{J}}
\newcommand{\cK}{\mathcal{K}}
\newcommand{\cL}{\mathcal{L}}
\newcommand{\cM}{\mathcal{M}}
\newcommand{\cN}{\mathcal{N}}
\newcommand{\cO}{\mathcal{O}}
\newcommand{\cP}{\mathcal{P}}
\newcommand{\cQ}{\mathcal{Q}}
\newcommand{\cR}{\mathcal{R}}
\newcommand{\cS}{\mathcal{S}}
\newcommand{\cT}{\mathcal{T}}
\newcommand{\cU}{\mathcal{U}}
\newcommand{\cV}{\mathcal{V}}
\newcommand{\cW}{\mathcal{W}}
\newcommand{\cX}{\mathcal{X}}
\newcommand{\cY}{\mathcal{Y}}
\newcommand{\cZ}{\mathcal{Z}}

\newcommand{\vA}{\mathbf{A}}
\newcommand{\vB}{\mathbf{B}}
\newcommand{\vC}{\mathbf{C}}
\newcommand{\vD}{\mathbf{D}}
\newcommand{\vE}{\mathbf{E}}
\newcommand{\vF}{\mathbf{F}}
\newcommand{\vG}{\mathbf{G}}
\newcommand{\vH}{\mathbf{H}}
\newcommand{\vI}{\mathbf{I}}
\newcommand{\vJ}{\mathbf{J}}
\newcommand{\vK}{\mathbf{K}}
\newcommand{\vL}{\mathbf{L}}
\newcommand{\vM}{\mathbf{M}}
\newcommand{\vN}{\mathbf{N}}
\newcommand{\vO}{\mathbf{O}}
\newcommand{\vP}{\mathbf{P}}
\newcommand{\vQ}{\mathbf{Q}}
\newcommand{\vR}{\mathbf{R}}
\newcommand{\vS}{\mathbf{S}}
\newcommand{\vT}{\mathbf{T}}
\newcommand{\vU}{\mathbf{U}}
\newcommand{\vV}{\mathbf{V}}
\newcommand{\vW}{\mathbf{W}}
\newcommand{\vX}{\mathbf{X}}
\newcommand{\vY}{\mathbf{Y}}
\newcommand{\vZ}{\mathbf{Z}}

\newcommand{\va}{\mathbf{a}}
\newcommand{\vb}{\mathbf{b}}
\newcommand{\vc}{\mathbf{c}}
\newcommand{\vd}{\mathbf{d}}
\newcommand{\ve}{\mathbf{e}}
\newcommand{\vf}{\mathbf{f}}
\newcommand{\vg}{\mathbf{g}}
\newcommand{\vh}{\mathbf{h}}
\newcommand{\vi}{\mathbf{i}}
\newcommand{\vj}{\mathbf{j}}
\newcommand{\vk}{\mathbf{k}}
\newcommand{\vl}{\mathbf{l}}
\newcommand{\vm}{\mathbf{m}}
\newcommand{\vn}{\mathbf{n}}
\newcommand{\vo}{\mathbf{o}}
\newcommand{\vp}{\mathbf{p}}
\newcommand{\vq}{\mathbf{q}}
\newcommand{\vr}{\mathbf{r}}
\newcommand{\Vs}{\mathbf{s}}
\newcommand{\vt}{\mathbf{t}}
\newcommand{\vu}{\mathbf{u}}
\newcommand{\vv}{\mathbf{v}}
\newcommand{\vw}{\mathbf{w}}
\newcommand{\vx}{\mathbf{x}}
\newcommand{\vy}{\mathbf{y}}
\newcommand{\vz}{\mathbf{z}}

\newcommand{\vone}{\mathbf{1}}
\newcommand{\vzero}{\mathbf{0}}

\newcommand{\valpha}{{\boldsymbol{\alpha}}}
\newcommand{\vbeta}{{\boldsymbol{\beta}}}
\newcommand{\vgamma}{{\boldsymbol{\gamma}}}
\newcommand{\vdelta}{{\boldsymbol{\delta}}}
\newcommand{\vepsilon}{{\boldsymbol{\epsilon}}}
\newcommand{\vzeta}{{\boldsymbol{\zeta}}}
\newcommand{\veta}{{\boldsymbol{\eta}}}
\newcommand{\vtheta}{{\boldsymbol{\theta}}}
\newcommand{\viota}{{\boldsymbol{\iota}}}
\newcommand{\vkappa}{{\boldsymbol{\kappa}}}
\newcommand{\vlambda}{{\boldsymbol{\lambda}}}
\newcommand{\vmu}{{\boldsymbol{\mu}}}
\newcommand{\vnu}{{\boldsymbol{\nu}}}
\newcommand{\vxi}{{\boldsymbol{\xi}}}
\newcommand{\vomikron}{{\boldsymbol{\omikron}}}
\newcommand{\vpi}{{\boldsymbol{\pi}}}
\newcommand{\vrho}{{\boldsymbol{\rho}}}
\newcommand{\vsigma}{{\boldsymbol{\sigma}}}
\newcommand{\vtau}{{\boldsymbol{\tau}}}
\newcommand{\vupsilon}{{\boldsymbol{\upsilon}}}
\newcommand{\vphi}{{\boldsymbol{\phi}}}
\newcommand{\vchi}{{\boldsymbol{\chi}}}
\newcommand{\vpsi}{{\boldsymbol{\psi}}}
\newcommand{\vomega}{{\boldsymbol{\omega}}}

\newcommand{\rLambda}{\mathrm{\Lambda}}
\newcommand{\rSigma}{\mathrm{\Sigma}}


\makeatletter
\newcommand{\vast}[1]{\bBigg@{#1}}
\makeatother


\newcommand{\pool}{\operatorname{pool}}


\def\roxf{$\cR$Oxf\xspace}
\def\rpar{$\cR$Par\xspace}
\def\roxfdist{$\cR$Oxf+$\cR$1M\xspace}
\def\rpardist{$\cR$Par+$\cR$1M\xspace}


\def \re{*}
\def \up{$\uparrow$}

\def \mac{MAC}
\def \gem{GeM}
\def \Mac{\mac\xspace}
\def \Gem{\gem\xspace}

\def \R{R}
\def \V{V}
\def \res{ResNet}
\def \vgg{VGG}
\def \Res{\res\xspace}
\def \Vgg{\vgg\xspace}

\def \W{W}
\def \D{D}
\def \Ddelf{\D$\dagger$\xspace}

\def \dsm{DSM}
\def \Dsm{\dsm\xspace}

\def \map{{\scriptsize mAP}}
\def \mpr{{\scriptsize mP@10}}
\def \Map{mAP\xspace}
\def \Mpr{mP@10\xspace}


\begin{abstract}
We propose a novel method of \emph{deep spatial matching} (DSM) for image retrieval. Initial ranking is based on image descriptors extracted from convolutional neural network activations by global pooling, as in recent state-of-the-art work. However, the same sparse 3D activation tensor is also approximated by a collection of local features. These local features are then robustly matched to approximate the optimal alignment of the tensors. This happens without any network modification, additional layers or training. No local feature detection happens on the original image. No local feature descriptors and no visual vocabulary are needed throughout the whole process.

We experimentally show that the proposed method achieves the state-of-the-art performance on standard benchmarks across different network architectures and different global pooling methods.
The highest gain in performance is achieved when diffusion on the nearest-neighbor graph of global descriptors is initiated from spatially verified images.

\end{abstract}

\section{Introduction}

\begin{figure}
\centering
\includegraphics[width=.8\columnwidth]{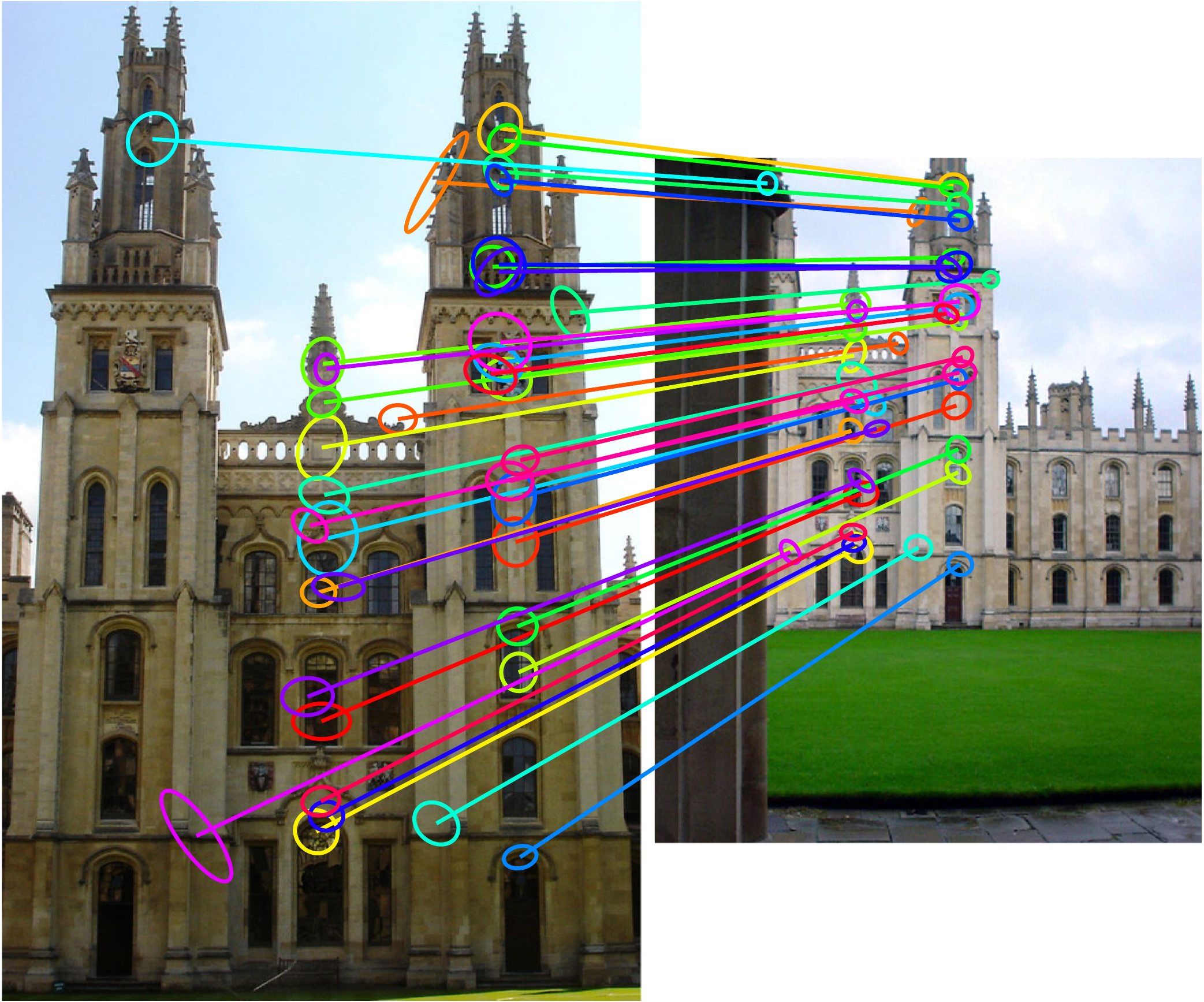}
\caption[caption]{Fast spatial matching~\cite{PCISZ07} finds a linear geometric transformation between two views of an object based on a local feature representation. This is used for spatial verification in large-scale image retrieval. Inlier correspondences shown, colored by visual word. What is the underlying representation?\setcounter{footnote}{0}\footnotemark\\\hspace{\textwidth}\parbox{\linewidth}{
\emph{\begin{enumerate}[topsep=0pt,itemsep=0pt]
	\item[(a)] SIFT~\cite{L99} descriptors on Hessian-affine~\cite{MS04} local features.
	\item[(b)] Descriptors on detected patches by an end-to-end differentiable pipeline using patch pair labels~\cite{yi2016lift}.
	\item[(c)] A subset of convolutional features at locations selected by an attention mechanism learned on image-level labels~\cite{noh2017largescale}.
	\item[(d)] Local maxima on each channel of a vanilla feature map. No vocabulary needed.
\end{enumerate}}}}
\label{fig:teaser}
\end{figure}

\footnotetext{\reflectbox{\rotatebox[origin=c]{180}{
\begin{minipage}[t]{\linewidth}
Answer: (d)~[this work].
\end{minipage}
}}}

Image and specific object retrieval is commonly addressed as large scale image matching: a query is matched against the database images and the final ranking is given by the matching score.
In the early retrieval days, methods based on local features were dominating~\cite{SZ03,NS06}. The matching score was first approximated by a similarity of bag of words~\cite{SZ03} or aggregated descriptors ~\cite{JDSP10}, and then re-ranked by efficient spatial verification~\cite{PCISZ07,PCM09}.

Recently, image retrieval is dominated by convolutional neural networks (CNNs)~\cite{GARL17,RTC18}.
Image representation is derived from the output of the CNN, which can be interpreted as a collection of 2D response maps of pattern detectors. The position of the response indicates the location of the pattern in the image, the size of the pattern is limited by the receptive field, and the value of the response indicates the confidence in the presence of the pattern.

Images of corresponding objects or object parts have similar response in all channels.
It is known that the image-to-image mapping can be recovered by correlating the response tensors of the two images~\cite{LoZD14,choy2016universal,rocco2018end}.

In general, the CNN activation tensor size depends on the number of channels and the image size. It is is too large to be stored, especially for large-scale applications.
To construct a descriptor of a fixed and reasonable size,
vectors obtained by global pooling are extracted instead,
for instance mean pooling~\cite{BL15}, max pooling~\cite{TSJ15}, generalized-mean pooling~\cite{RTC18}, and others~\cite{KMO16,TSJ15}. If the CNN-response tensors are matching, the statistics obtained after the global pooling should be matching too.

Global pooling not only reduces the size of the descriptor, but also injects view-point invariance. In fact, the global pooling is, similarly as bag of features, invariant to a very broad class of transformations. Thus, some information, namely geometric consistency, is lost.

In this work we introduce a very simple way of extracting from the CNN activations a representation that is suitable for geometric verification, which we apply to re-ranking.
Ideally, one would estimate the geometric transformation to align the activation tensors and compare. Nevertheless, as stated previously,
this would be impractical.
We propose to approximate this process, exploiting two properties of the activations: high values are more important and the activations are sparse. Therefore each channel can be well approximated by a small number of extremal regions.

After discussing related work in section~\ref{sec:related}, we develop our method, called \emph{deep spatial matching} (\Dsm), in section~\ref{sec:method}. Experimental results are reported in section~\ref{sec:exp} and conclusions are drawn in section~\ref{sec:discussion}.

\section{Related work}
\label{sec:related}

Shortly after the popularization of AlexNet and the illustration of image retrieval using the output vector of its last \emph{fully connected} layer~\cite{KrSH12}, it was found that convolutional layers possessed much more discriminative power and were better adapted to new domains~\cite{BSCL14}. However, just flattening the 3D \emph{convolutional activation} tensor into a vector yields a non-invariant representation. The next obvious attempt was to split images into patches, apply \emph{spatial max-pooling} and match them exhaustively pairwise, which could beat conventional pipelines~\cite{TAJ13} for the first time, but is expensive~\cite{RSMC14}. Is was then shown more efficient to apply regional max-pooling on a single convolutional activation of the entire image~\cite{TSJ15}. Combined with integral image computation, \cite{TSJ15}~also allowed fast sliding window-style spatial matching, still requiring to store a tensor of the same size as the entire convolutional activation tensor.

Network \emph{fine-tuning} of globally pooled representations like MAC and R-MAC~\cite{TSJ15} using metric learning loss functions for the retrieval task followed, giving state of the art performance~\cite{GARL17,RTC18}. The power of CNN representations of one or very few regional descriptors per image allowed reducing image retrieval to nearest neighbor search and extending previous query expansion~\cite{CPSIZ07,ToJe14} into efficient online exploration of the entire nearest neighbor graph of the dataset by \emph{diffusion}~\cite{ITA+16}. This progress nearly solved previous benchmarks and necessitated revisiting them~\cite{RIT+18}. The main drawback of these compact representations is that they are not compatible with spatial verification, which would ensure accuracy of the top ranking results as was the case with conventional
representations~\cite{PCISZ07,ToAv11}. In fact, given enough memory, such representations are still the state of the art~\cite{RIT+18}.

Most notable in the latter benchmark was the performance of \emph{deep local features} (DELF)~\cite{noh2017largescale}, which combined the power of CNN features with the conventional pipeline of hundreds of local descriptors per image, followed by encoding against a vocabulary and search by inverted files. The DELF apprach does allow spatial verification at the cost of more memory and incompatibility with global representations, which on the other hand, allow nearest neighbor search. In this work, we attempt to reduce this gap by introducing a new representation that
encodes geometric information allowing spatial verification, yet it has a trivial relation to the global representation used for nearest neighbor search.

At this point, it is worth looking at the geometric alignment of two views shown in Figure~\ref{fig:teaser} and reflecting on what could be the underlying representation and what would be the advantages of each choice. In terms of geometric correspondence, most recent efforts have focused on either dense registration~\cite{LoZD14,choy2016universal,rocco2018end,RCA+18}, which would not apply to retrieval due to the storage limitation, or imitating conventional pipelines~\cite{L99,MS04}. In the latter case, two dominating paradigms are \emph{detect-then-describe}~\cite{yi2016lift} and \emph{describe-then-detect}~\cite{noh2017largescale}, both of which result in a large set of visual descriptors. We break this dilemma by ``reading off'' information directly from feature maps.
\section{Deep spatial matching}
\label{sec:method}

We begin by motivating our approach, and then present the proposed architecture, followed by the main ideas, including feature detection and representation from CNN activations, spatial matching and re-ranking.

\subsection{Motivation}
\label{sec:idea}

\begin{figure*}
\centering
\begin{tabular}{cccc}
	\includegraphics[height=2.6cm]{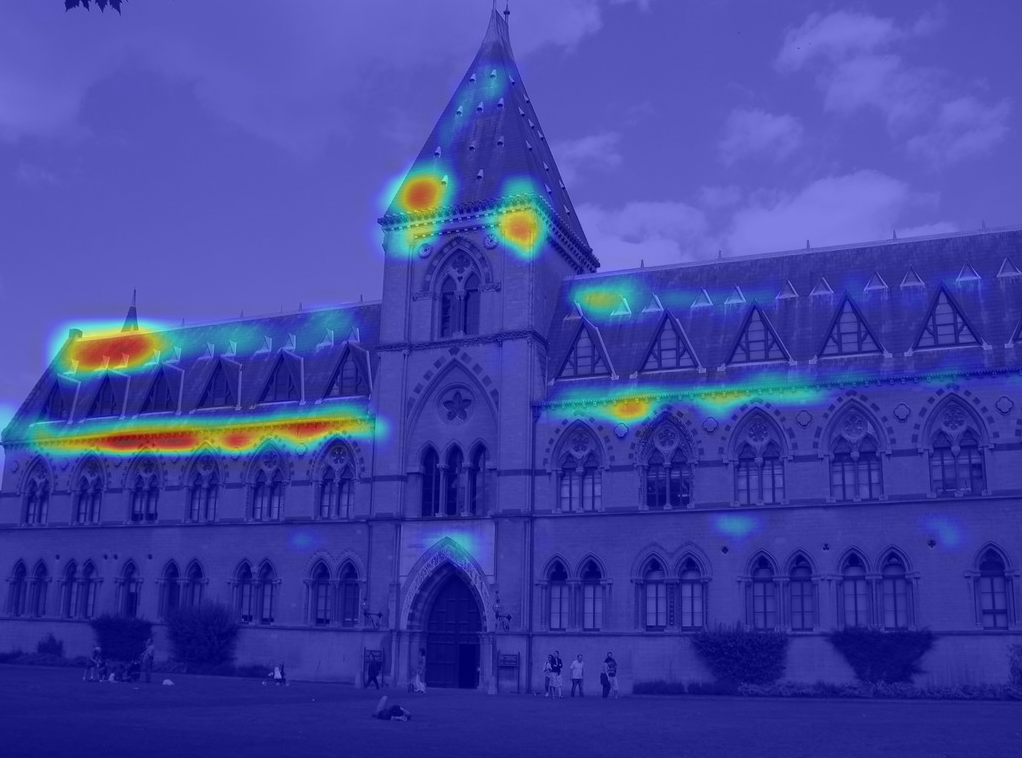} &
	\includegraphics[height=2.6cm]{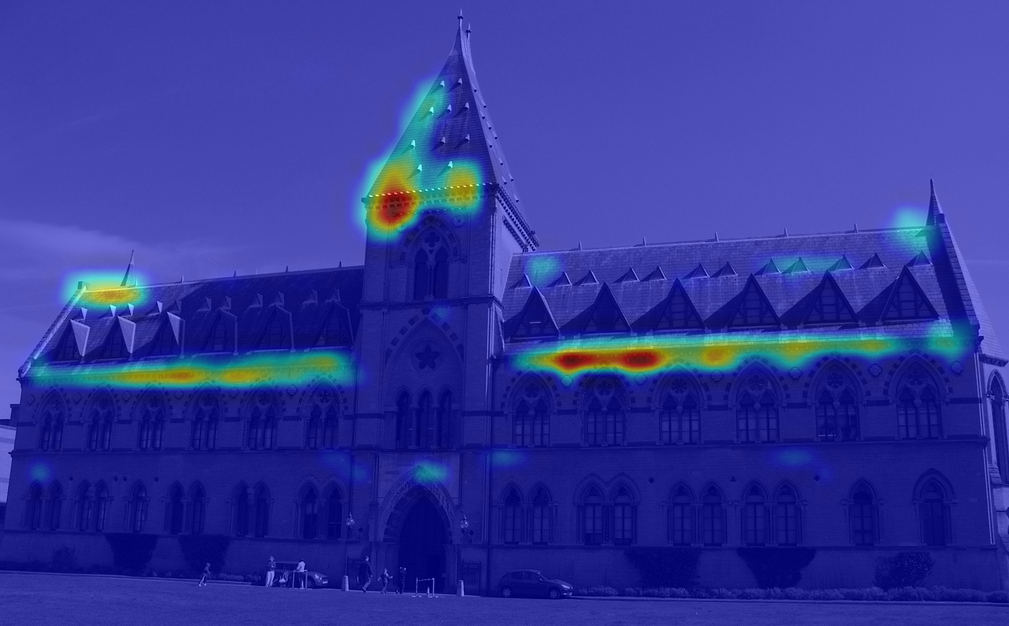} &
	\includegraphics[height=2.6cm]{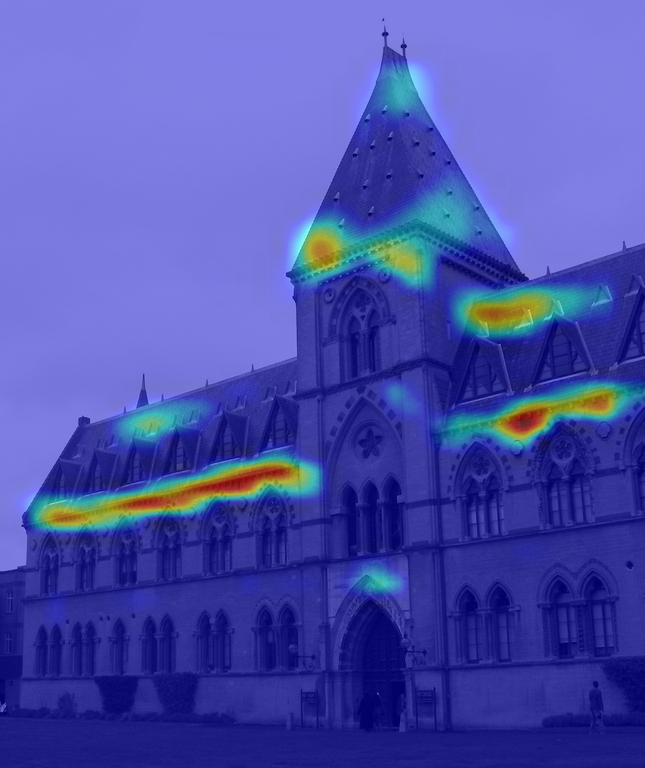} &
	\includegraphics[height=2.6cm]{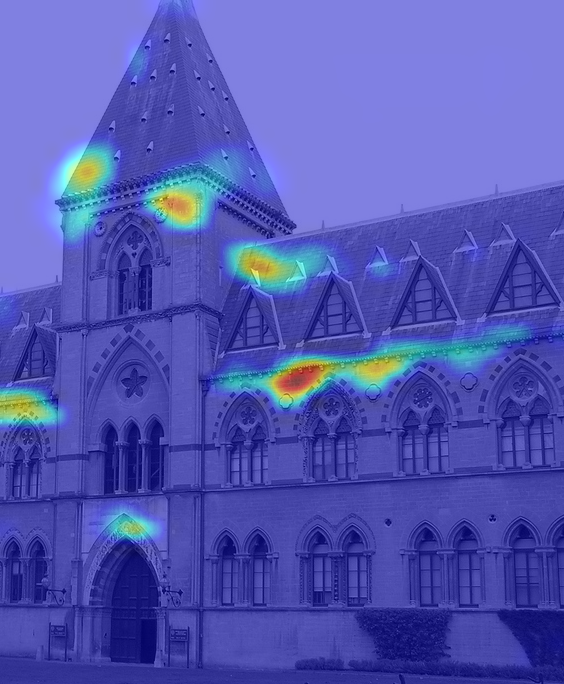} \\

	\includegraphics[height=2.6cm]{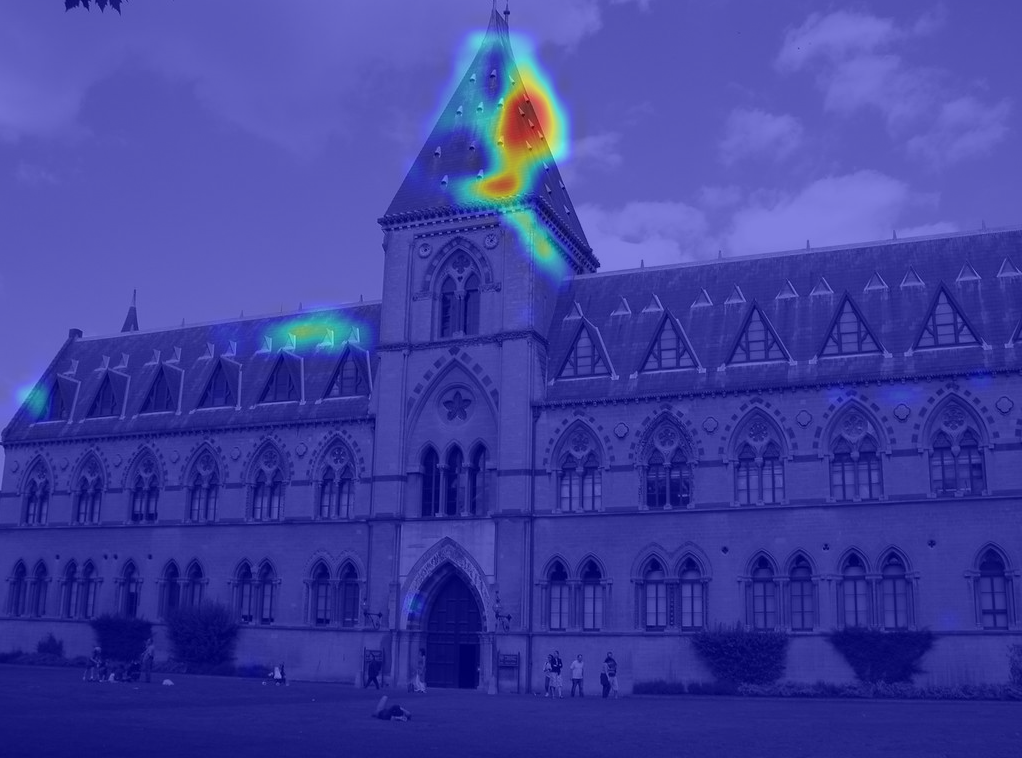} &
	\includegraphics[height=2.6cm]{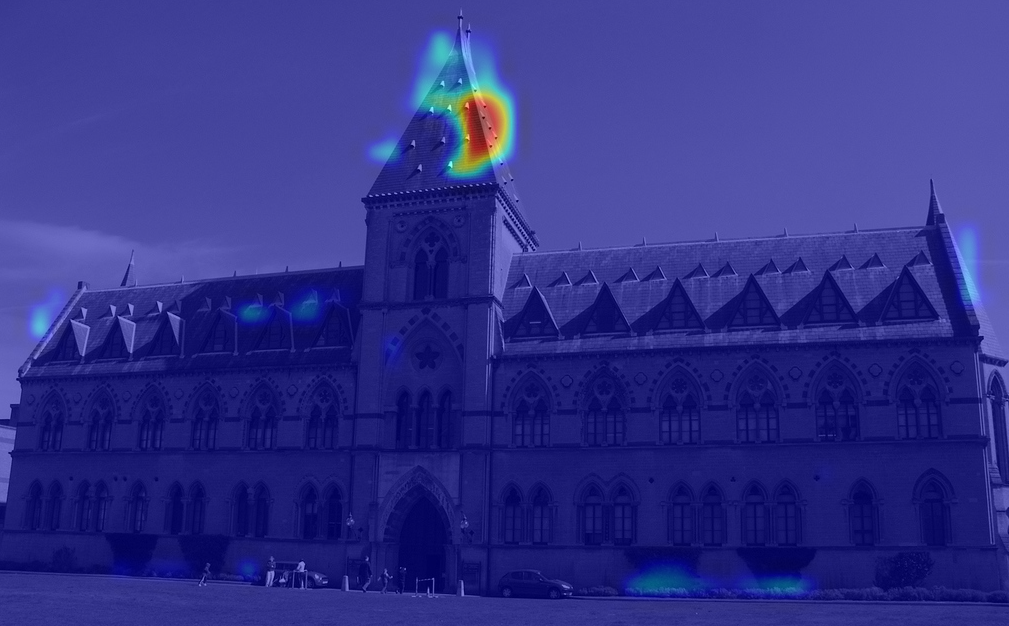} &
	\includegraphics[height=2.6cm]{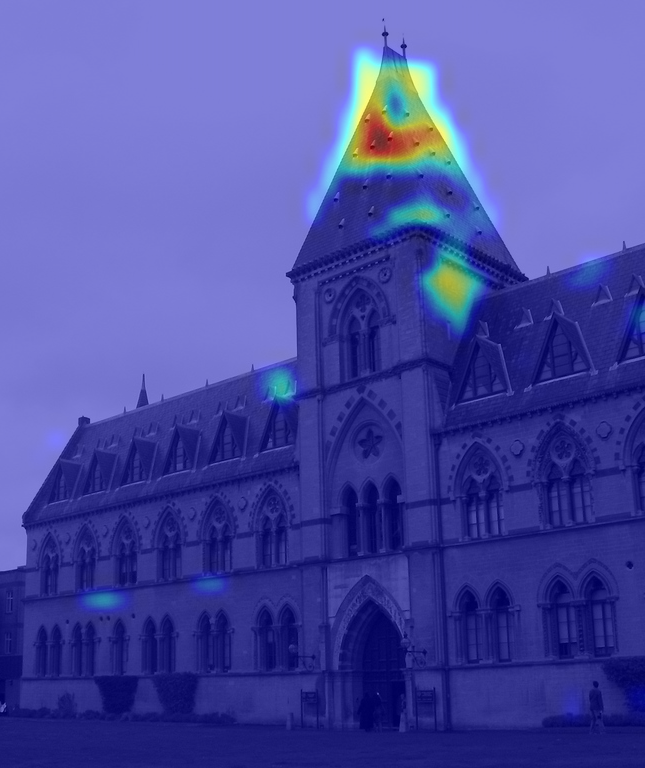} &
	\includegraphics[height=2.6cm]{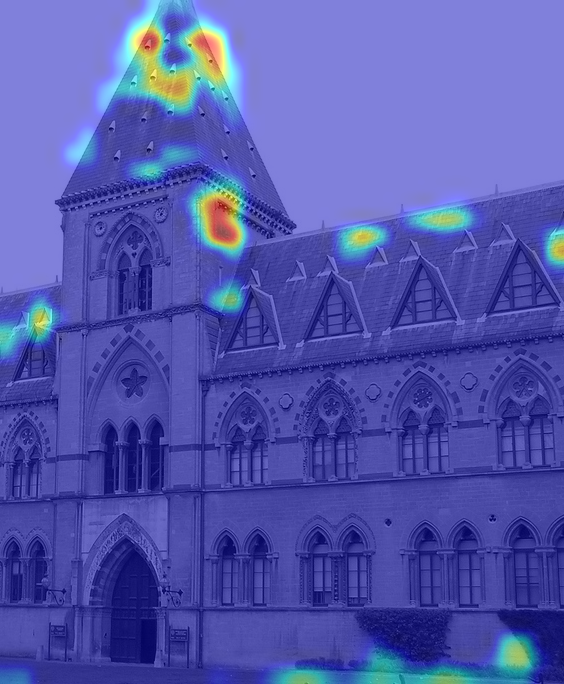}  \\
\end{tabular}
\caption{Four views (columns) of the Museum of natural history in the \roxf dataset, overlaid with two different feature maps (rows) of the last convolutional layer of the VGG16~\cite{SZ14} network. The filter kernel in each channel is responding to similar image structures in all images. All activations are naturally sparse and nonzero responses agree in both location and local shape between all images.}
\label{fig:fm}
\end{figure*}

Given a convolutional neural network ending in global average pooling, objects of a given class can be localized by \emph{class activation maps} (CAM)~\cite{ZKL+16}, even if the network has only been trained for classification on image-level labels. These maps are linear combinations of individual feature maps (channels) of the last convolutional layer. \alert{Grad-CAM~\cite{SCDVPB17} generalizes this idea to any network architecture and allows visualization at any layer by a similar linear combination on the gradient signal instead.} Without any class semantics, another linear combination produces a saliency map used for spatial pooling in \emph{cross-dimensional weighting} (CroW)~\cite{KMO16}. The latter weighs channels according to \emph{sparsity}, but in all cases the linear combinations only provide coarse localization of objects of a given class or class-agnostic salient regions.

Experiments in~~\cite{TSJ15} have shown \emph{max-pooling} of convolutional activations (MAC) to be
superior to other spatial pooling schemes, at least for image retrieval.
This can be connected to the sparsity of the activations. More interestingly, looking at the positions of the maxima in channels contributing most to image similarities, one can readily identify correspondences between two images~\cite{TSJ15}. The same has been observed in person re-identification~\cite{almazan2018re}. Later, \emph{generalized mean pooling} (GeM)~\cite{RTC18} was shown to outperform max-pooling.
This can be attributed to the fact that it allows for more than one locations contributing to the representation, while still being more selective than average pooling.

Following the above observations, we investigate the responses of the last convolutional layer of VGG on several matching images of the \roxf dataset. This time we do not limit ourselves to the channels that are contributing most to image similarity (assuming \eg global max-pooling and cosine similarity), but we rather observe all channels. We find out that, as illustrated in Figure~\ref{fig:fm}, for two example channels, in most cases the responses to all images are not just sparse but consistent too: the filters respond to the same structures in the images, and there are responses at consistent locations with consistent local shape. The responses exhibit translation and scale covariance to some extent.
The deep spatial matching proposed in this work is motivated by the following ideas.

\emph{Instead of just reducing each channel to a single scalar, why not keep all the peaks of the responses in each channel along with geometric information (coordinates and local shape)? Instead of attaching an entire descriptor to each such geometric entity, why not just attach the channel it was extracted from, as if it was a visual word?}

We propose a method in-between two commonly used approaches, taking the best of the two worlds.
One is conventional representations of thousands of local features per image, each with its own descriptor, suitable for inverted files and spatial verification. The other relies on a single global or few regional descriptors per image, leading to compact storage, efficient nearest neighbor search, and graph-based re-ranking.
The proposed approach is applicable to any network fine-tuned for retrieval, without requiring any network adaptation, even without any training. It needs no vocabulary and it is trivially related to the global descriptors that dominate the state of the art.

\subsection{Method overview}
\label{sec:architecture}

\begin{figure*}
\centering
\extfig{arch}{\begin{tikzpicture}[
	every node/.append style={font=\scriptsize},
	fun/.style={semithick,black},
	diff/.style={semithick,black},
]
	\newcommand{\inliera}{
		\xell[ell1]{(4,0,-.3)}{.3,.5}{0}
		\xell[ell1]{(4,1.3,-.3)}{.6,.3}{0}
		\xell[ell1]{(4,2.8,0)}{.7,.4}{60}
		\xell[ell1]{(4,-.5,2)}{.25,.25}{0}
	}
	\newcommand{\inlierb}{
		\xell[ell2]{(4,-.5,-1)}{.3,.5}{0}
		\xell[ell2]{(4,.8,-1)}{.5,.3}{20}
		\xell[ell2]{(4,2.8,-.7)}{.7,.3}{80}
		\xell[ell2]{(4,-1,1.6)}{.2,.2}{0}
	}
	\matrix[
		row sep=0pt,column sep=0pt,cells={scale=.18},
		para/.style={draw,depth=.25},
		ell1/.style={draw,blue},
		ell2/.style={draw,red},
		ar/.style={above right=2pt and 2pt},
		br/.style={below right=2pt and 2pt},
	]
	{
		\node at(0,0) {input image}; \&
		\path (0,0)--(6,0); \&
		\node at(2,0) {feature map}; \&
		\path (0,0)--(8,0); \&
		\node at(2,0) {local features}; \&
		\path (0,0)--(3,0); \&
		\node at(2,0) {inliers}; \&
		\path (0,0)--(3,0);
		\\
		\begin{scope}[scale=1.2]
			\paraim[para]{0,-5,-4}{.5,10,8}{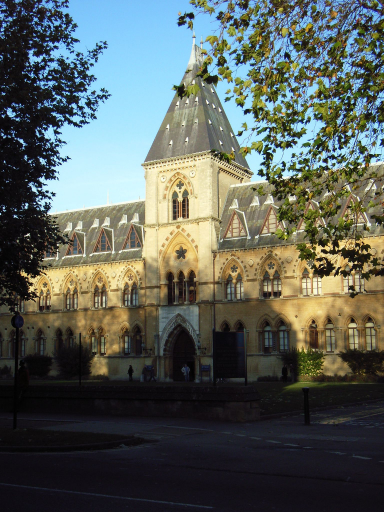}
		\end{scope}
		\node at(3,-7) {$x_1$};
		\node (im1-r) at(4,0) {};
		\&\&
		\paraim[para]{0,-5,-4}{4,10,8}{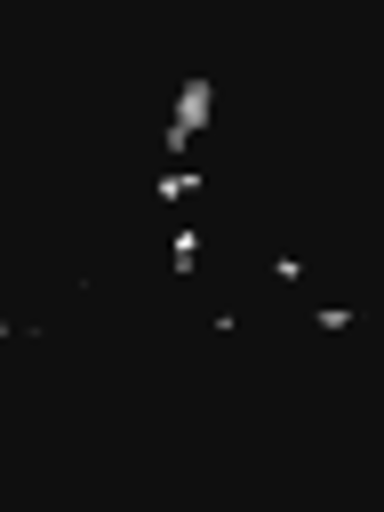}
		\node at(6,-6) {$A_1$};
		\node (fm1-l) at(-2.5,0) {};
		\node (fm1-r) at(6.5,0) {};
		\node (fm1-t) at(2,4.5) {};
		\&\&
		\para[para]{0,-5,-4}{4,10,8}
		\inliera
		\xell[ell1]{(4,-1.8,3)}{.35,.25}{0}
		\xell[ell1]{(4,-2,1)}{.25,.25}{0}
		\node at(6,-6) {$\cP_1$};
		\node (lf1-l) at(-2.5,0) {};
		\node (lf1-b) at(0,-6) {};
		\\[-24pt]
		\&\&\&\&
		\node[node,fun] (g) at(2,0) {$g$};
		\node[left=6pt,fun] at(g) {spatial\\matching}; \&\&
		\para[para]{0,-4,-4}{4,8,8}
		\inliera
		\begin{scope}[shift={(0,-.2,-2)},scale=.9]
			\inlierb
		\end{scope}
		\node at(-2,5) {$\cM$};
		\node (in-l) at(-2.5,0) {};
		\node (in-r) at(7,0) {};
		\node (in-t) at(2,4) {};
		\node (in-b) at(2,-5) {};
		\&\&
		\node[node,fun] (si) at(0,0) {$s$};
		\node[br,fun] at(si) {similarity\\function};
		\draw[->,fun] (si)--(5,0);
		\\[-20pt]
		\begin{scope}[scale=1.2]
			\paraim[para]{0,-4,-6}{.5,8,12}{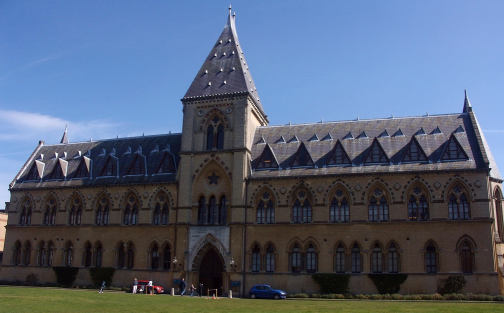}
		\end{scope}
		\node at(-4,5.5) {$x_2$};
		\node (im2-r) at(5.5,0) {};
		\&\&
		\paraim[para]{0,-4,-6}{4,8,12}{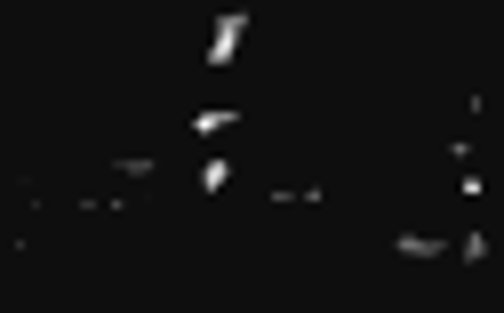}
		\node at(-3,5) {$A_2$};
		\node (fm2-l) at(-4,0) {};
		\node (fm2-r) at(8,0) {};
		\node (fm2-b) at(3,-4) {};
		\&\&
		\para[para]{0,-4,-6}{4,8,12}
		\inlierb
		\xell[ell2]{(4,-1,.7)}{.3,.2}{0}
		\xell[ell2]{(4,-2.2,3.8)}{.4,.25}{0}
		\xell[ell2]{(4,-2.2,5.2)}{.25,.4}{0}
		\xell[ell2]{(4,-.5,5.2)}{.25,.25}{0}
		\xell[ell2]{(4,.5,5)}{.2,.2}{0}
		\xell[ell2]{(4,1.6,5.2)}{.2,.3}{0}
		\xell[ell2]{(4,-.2,-2.7)}{.4,.25}{0}
		\xell[ell2]{(4,-1.2,-3.1)}{.2,.2}{0}
		\xell[ell2]{(4,-1.2,-4)}{.25,.2}{0}
		\node at(-3,5) {$\cP_2$};
		\node (lf2-l) at(-4,0) {};
		\node (lf2-t) at(0,4) {};
		\\
	};
	\draw[->,diff] (im1-r)--node[below]{CNN} node[above]{$f$} (fm1-l);
	\draw[->,diff] (im2-r)--node[below]{CNN} node[above]{$f$} (fm2-l);
	\draw[->,fun] (fm1-r)--node[below]{feature\\detection} node[above]{$d$} (lf1-l);
	\draw[->,fun] (fm2-r)--node[below]{feature\\detection} node[above]{$d$} (lf2-l);
	\draw[->,fun] let \p1=(lf1-b),\p2=(g) in (\x2,\y1)--(g.north);
	\draw[->,fun] let \p1=(lf2-t),\p2=(g) in (\x2,\y1)--(g.south);
	\draw[->,fun] (g.east)--(in-l);
	\draw[->,fun] let \p1=(in-r),\p2=(si) in (\x1,\y2)--(si.west);
\end{tikzpicture}}
\caption{\emph{Deep spatial matching} (\Dsm) network architecture.
Two input images $x_1,x_2$ are mapped by network $f$ to feature tensors $A_1,A_2$ respectively.
Sparse \emph{local features} $\cP_1,\cP_2$ extracted by \emph{detector} $d$ undergo \emph{spatial matching} $g$, resulting in a collection of inliers $\cM$.
Similarity function $s$ applies to this collection.
Local features are detected and matched independently per channel, with channels playing the role of \emph{visual words}. This takes place without
any additional learning and without adapting the backbone network.
In retrieval, only local features $\cP_1,\cP_2$ are stored and $g$ applies directly at re-ranking.
}
\label{fig:arch}
\end{figure*}
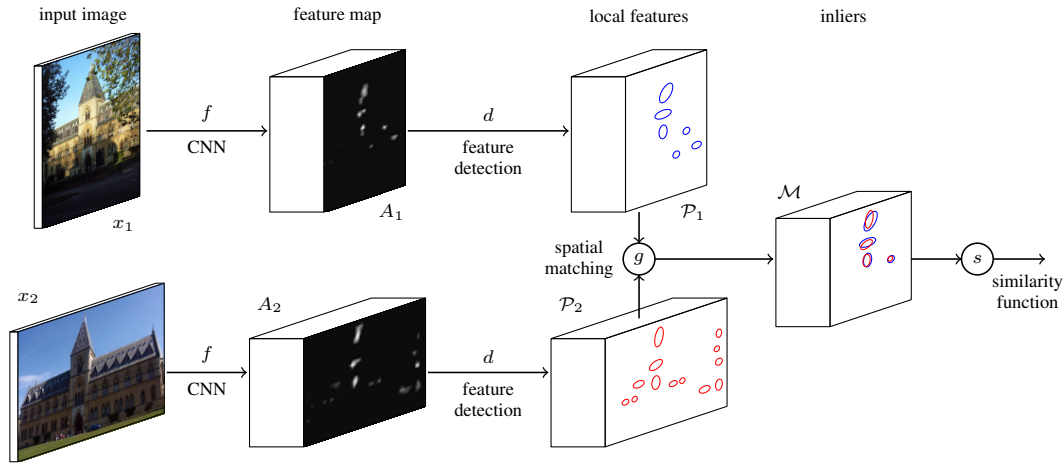

The preceding ideas give rise to the \emph{deep spatial matching} (\Dsm) network architecture that we introduce in this work, illustrated in Figure~\ref{fig:arch}. We consider a fully convolutional backbone network architecture that maintains as much as possible spatial resolution. We denote by $f$ the \emph{network function} that maps an input image to the feature tensor of the last convolutional layer. We assume that the backbone network $f$, when followed by a pooling mechanism \eg MAC~\cite{TSJ15} or GeM~\cite{RTC18}, extracts a global descriptor that is used \eg for retrieval~\cite{GARL17,RTC18}.

As shown in Figure~\ref{fig:arch},
two input images $x_1,x_2$ are processed by a network into
3-dimensional \emph{feature tensors} $A_1 \defn f(x_1), A_2 \defn f(x_2)$ where $A_i \in \real^{w_i \times h_i \times k}$, $w_i \times h_i$ is the spatial resolution of $A_i$ for $i=1,2$ and $k$ is the number of channels (features). Using the two feature tensors is standard practice in image registration~\cite{LoZD14,choy2016universal}, optical flow~\cite{dosovitskiy2015flownet} or semantic alignment~\cite{kim2017fcss,rocco2018end}, but here we use an entirely different way of working with the tensors.

In particular, similarly to local feature detection from a single feature tensor~\cite{noh2017largescale}, most registration/flow/alignment methods see a feature tensor $A \in \real^{w \times h \times k}$ as a $w \times h$ array of $k$-dimensional vector descriptors. Then, given two feature tensors, most consider the correlation of the two 2-dimensional arrays, seeking dense correspondences. By contrast, from each feature tensor $A_1,A_2$ we extract a sparse collection of \emph{local features} $\cP_1 \defn d(A_1), \cP_2 \defn d(A_2)$ respectively. The feature detector $d$, discussed in section~\ref{sec:detection}, operates independently per channel and each local feature collection $\cP$ is a list of sets, one per channel. Local features are represented as discussed in section~\ref{sec:repr}.

Then, the two local feature collections $\cP_1,\cP_2$ undergo \emph{spatial matching}, denoted as $g$ and discussed in section~\ref{sec:match}, returning a collection of inliers $\cM$ and a geometric transformation $T$. We fit a linear motion model to a collection of tentative \emph{correspondences}, \ie, pairs of local features from the two images, which are formed again independently per channel. This implicitly assumes that the ``appearance'' of each local feature is \emph{quantized} according to channel where it was detected, hence channels play the role of \emph{visual words}, without any descriptor vectors ever being computed. The output collection of \emph{inlier} correspondences $\cM$ is again given as a list of sets, one per channel.
Finally, \emph{similarity function} $s$ applies to $\cM$.

The entire feature detection and matching mechanism takes place without adapting the backbone network in any way and without any additional learning.
When applied to image retrieval, this architecture assumes that local features have been precomputed and are the representation of the database images, that is, feature tensors are discarded. Based on this representation, spatial matching $g$ applies directly for geometric verification and \emph{re-ranking}.

\subsection{Local feature detection}
\label{sec:detection}

To detect local features in each feature channel we use \emph{maximally stable extremal regions} (MSER) by Matas \etal~\cite{MCMP02}. MSERs are defined over a 2-dimensional input, in our case over
feature map $A^{(j)}$
of feature tensor $A$ independently for each channel $j=1,\dots,k$.
The extractor finds continuous regions $R$ with all interior points having strictly higher response value than neighboring outer points. Regions satisfying a stability criterion~\cite{MCMP02} and passing location non-maxima suppression are selected. These features are appropriate for regions of arbitrary shape, including localized peaks, blobs, elongated or even nested regions.

When MSERs are used as image features, the response value is
either the image intesity (MSER$^+$) or the reverse intensity (MSER$^-$). In our case, only regions of high CNN activations in sparse feature maps are of interest, and hence only one type of MSERs are extracted directly over the feature map responses.

\begin{figure}
\centering
\begin{tabular}{cc}
	\includegraphics[height=3.4cm]{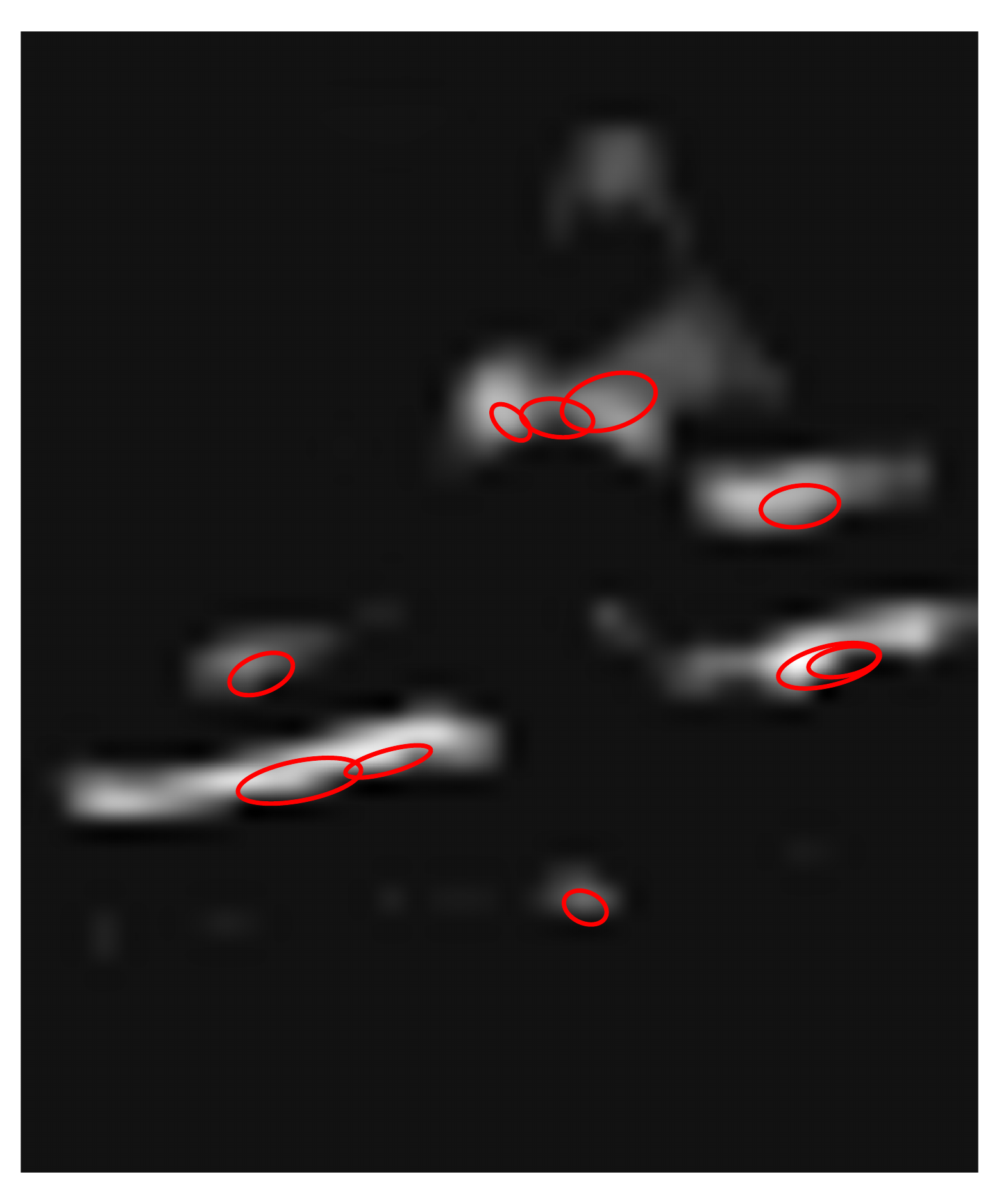} &
	\includegraphics[height=3.4cm]{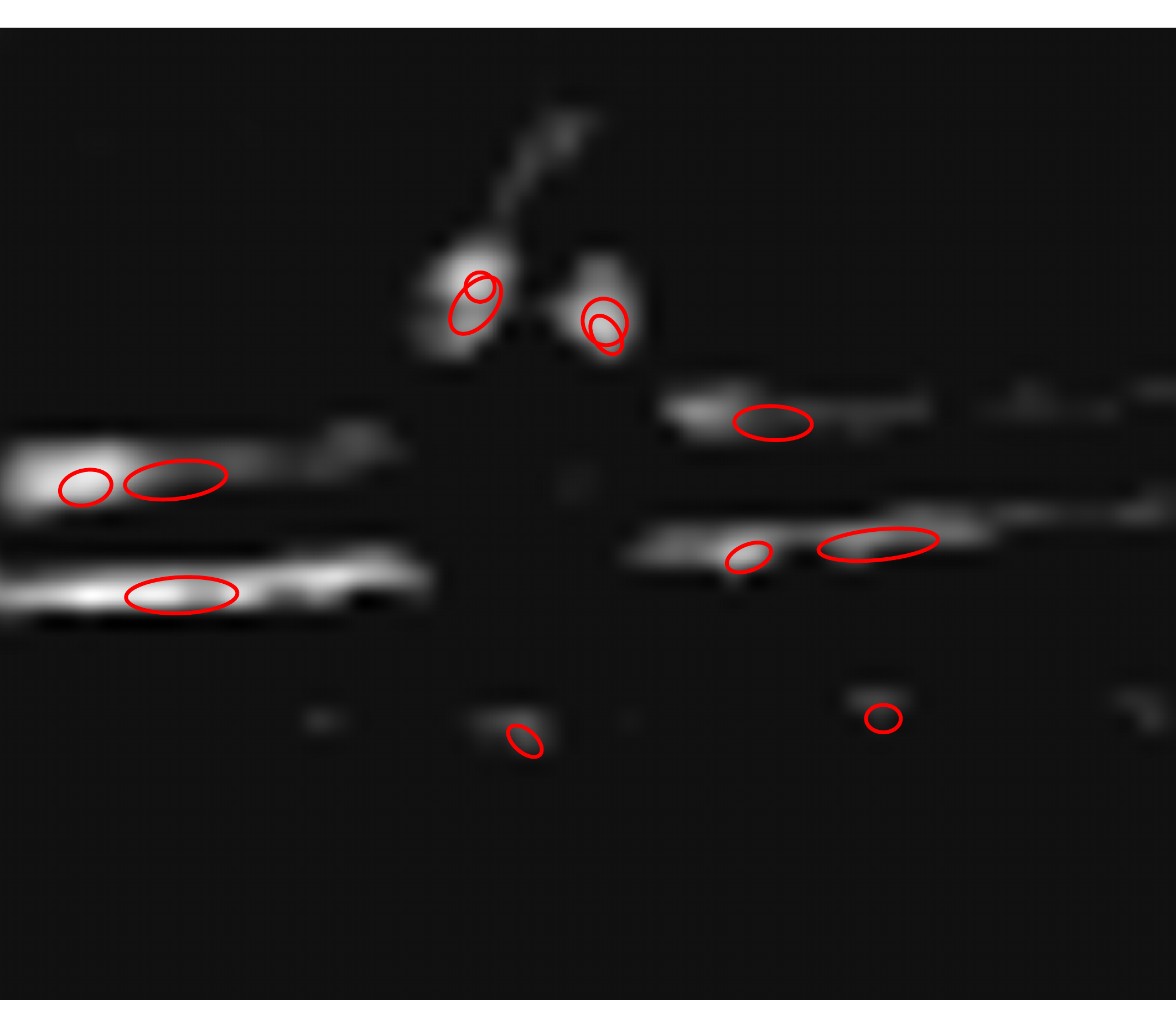} \\
\end{tabular}
\caption{Feature maps from one channel for two different views of a building in the \roxf dataset. Ellipses are fitted to the local features detected by MSER.}
\label{fig:fm_ell}
\end{figure}

\begin{figure*}
\begin{tabular}{cccc}
	\includegraphics[height=3.2cm]{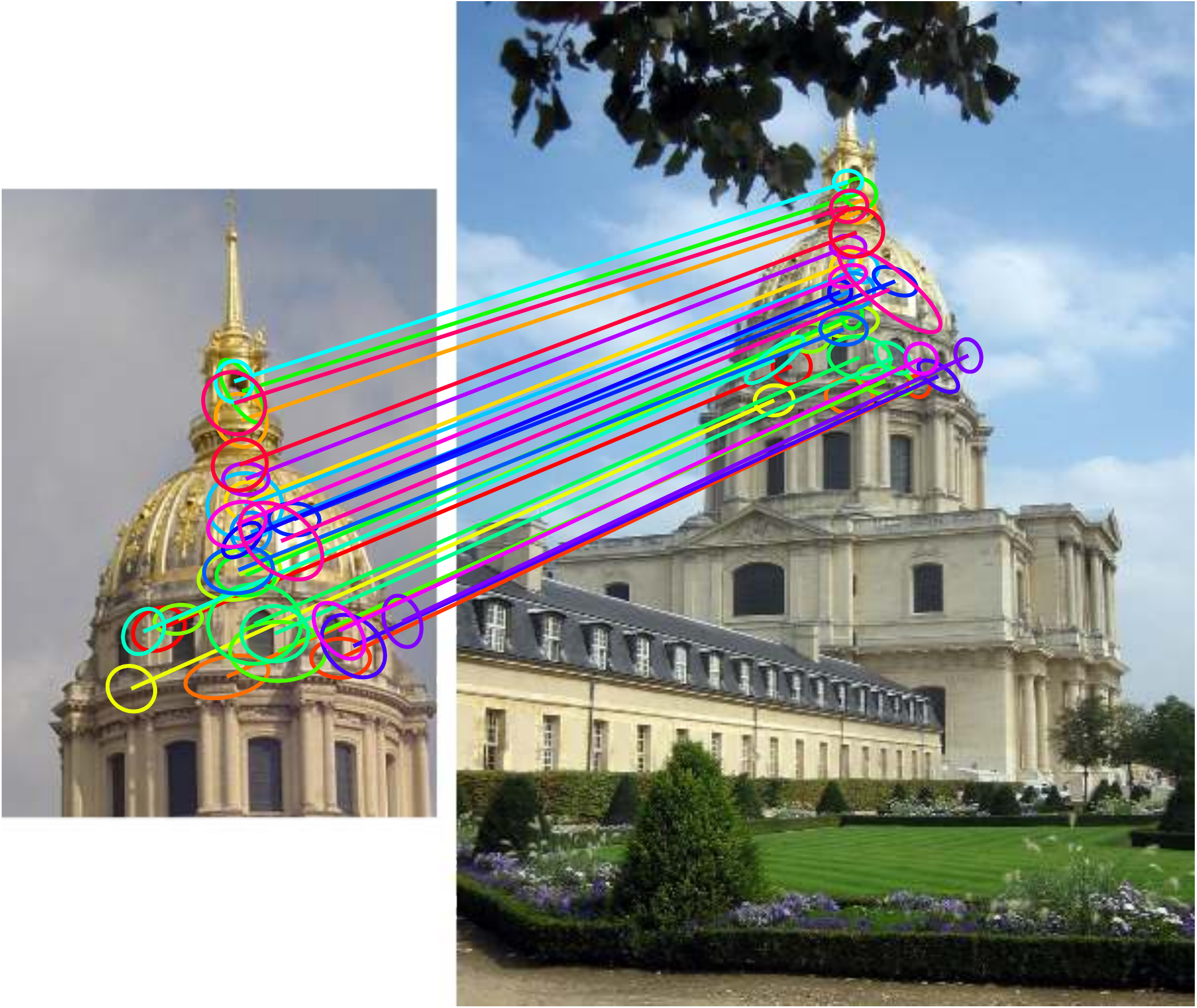} &
	\includegraphics[height=3.2cm]{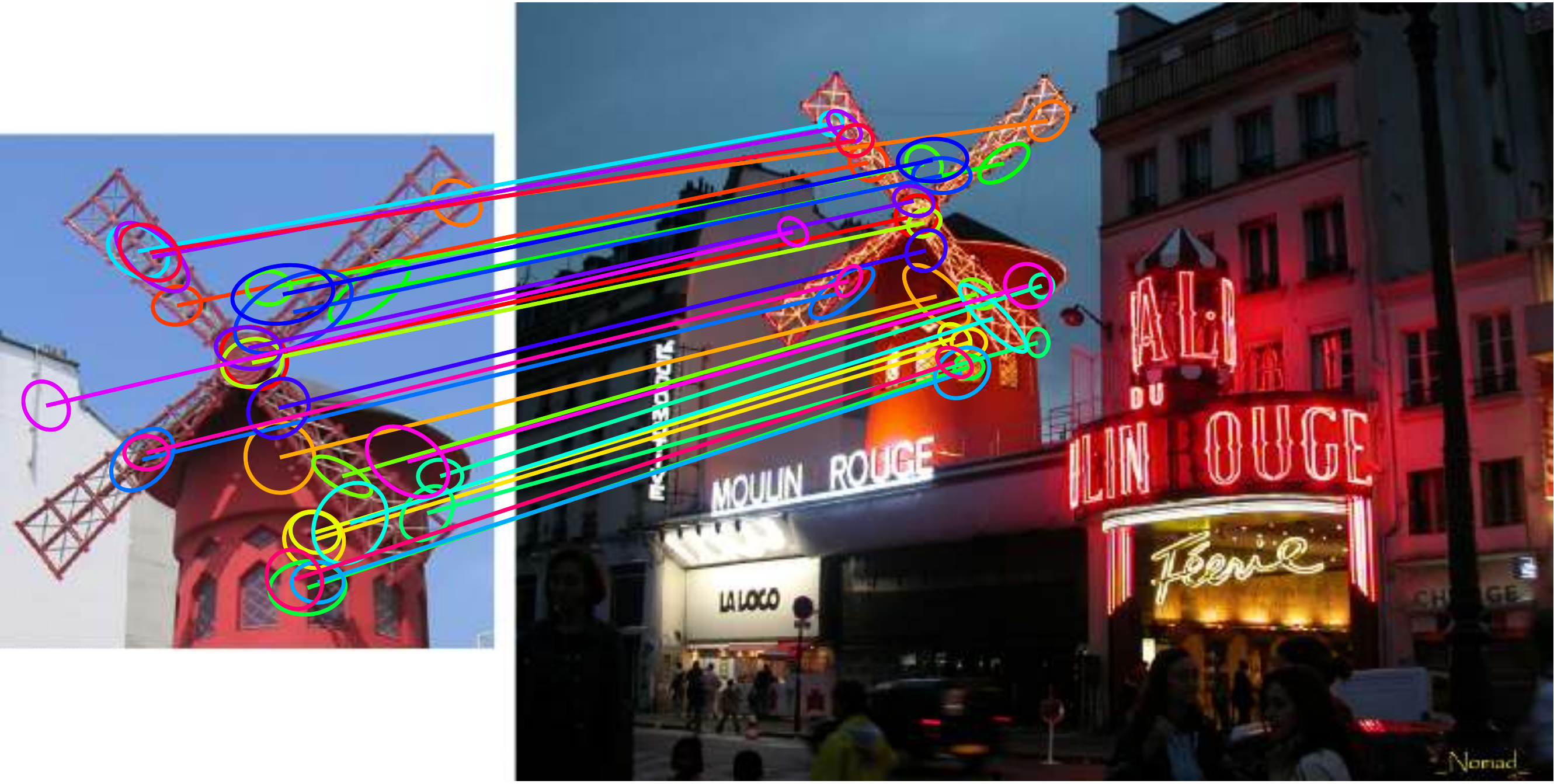} &
	\includegraphics[height=3.2cm]{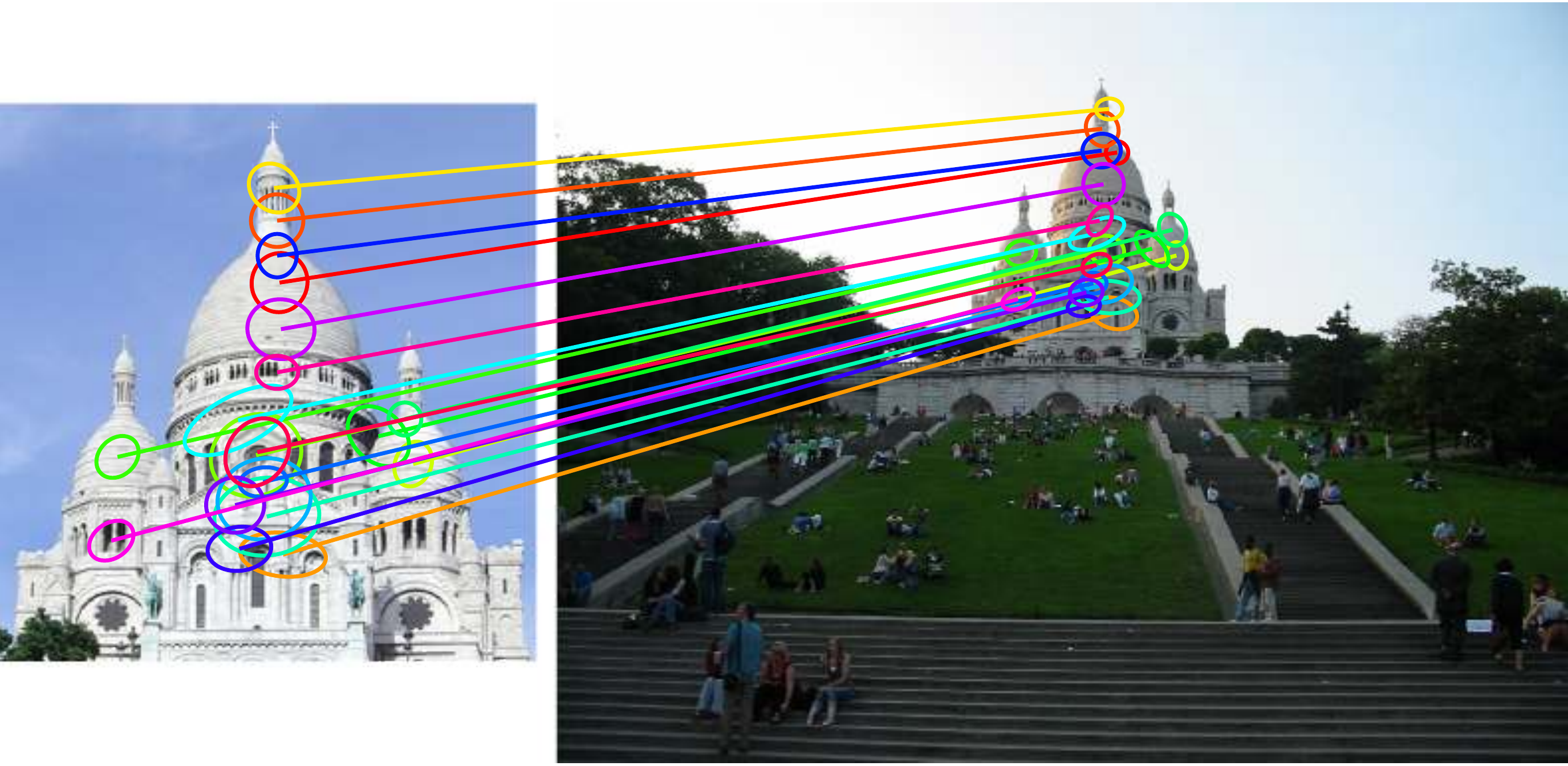}
	\\

	\includegraphics[height=3.2cm]{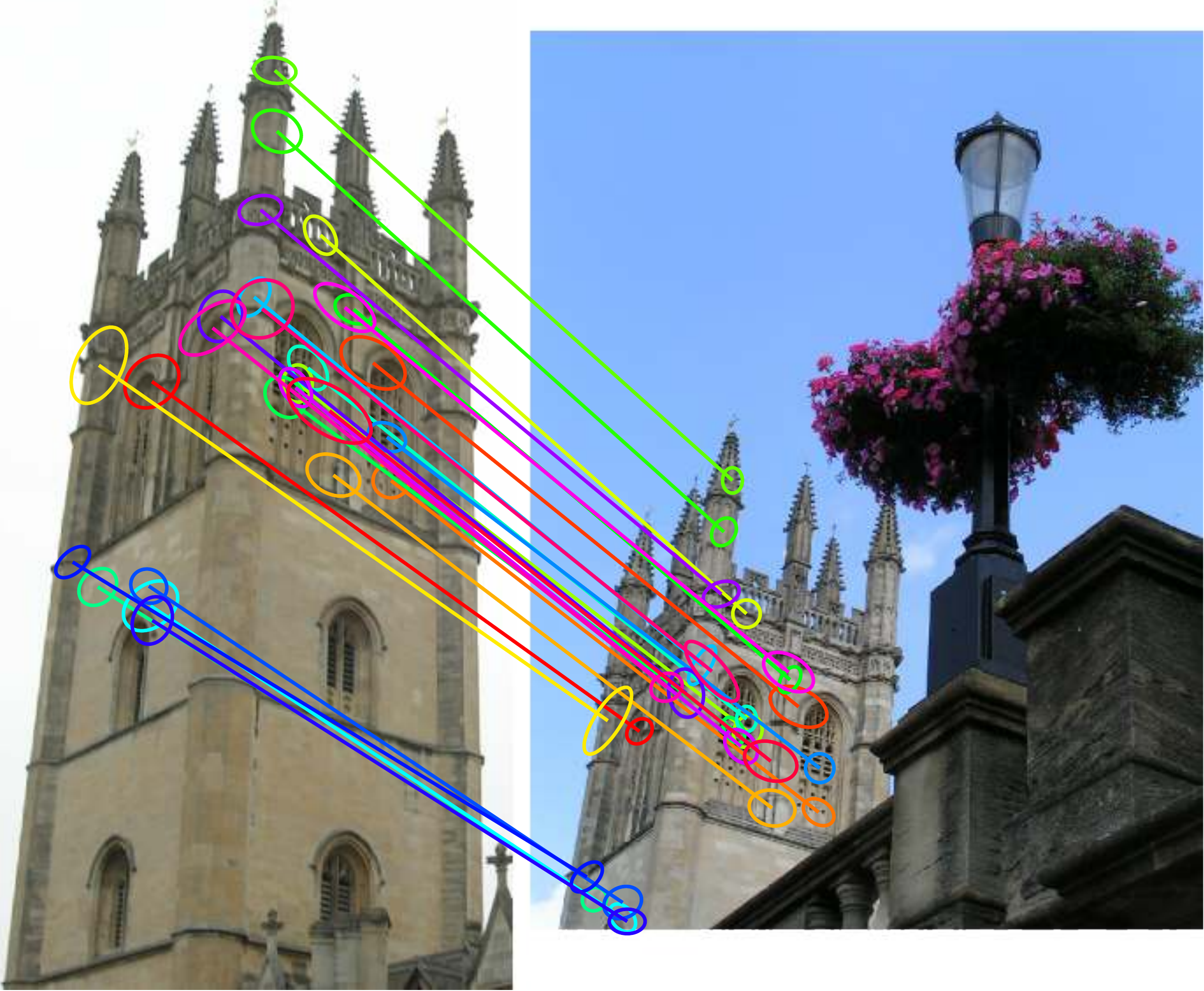} &
	\includegraphics[height=3.2cm]{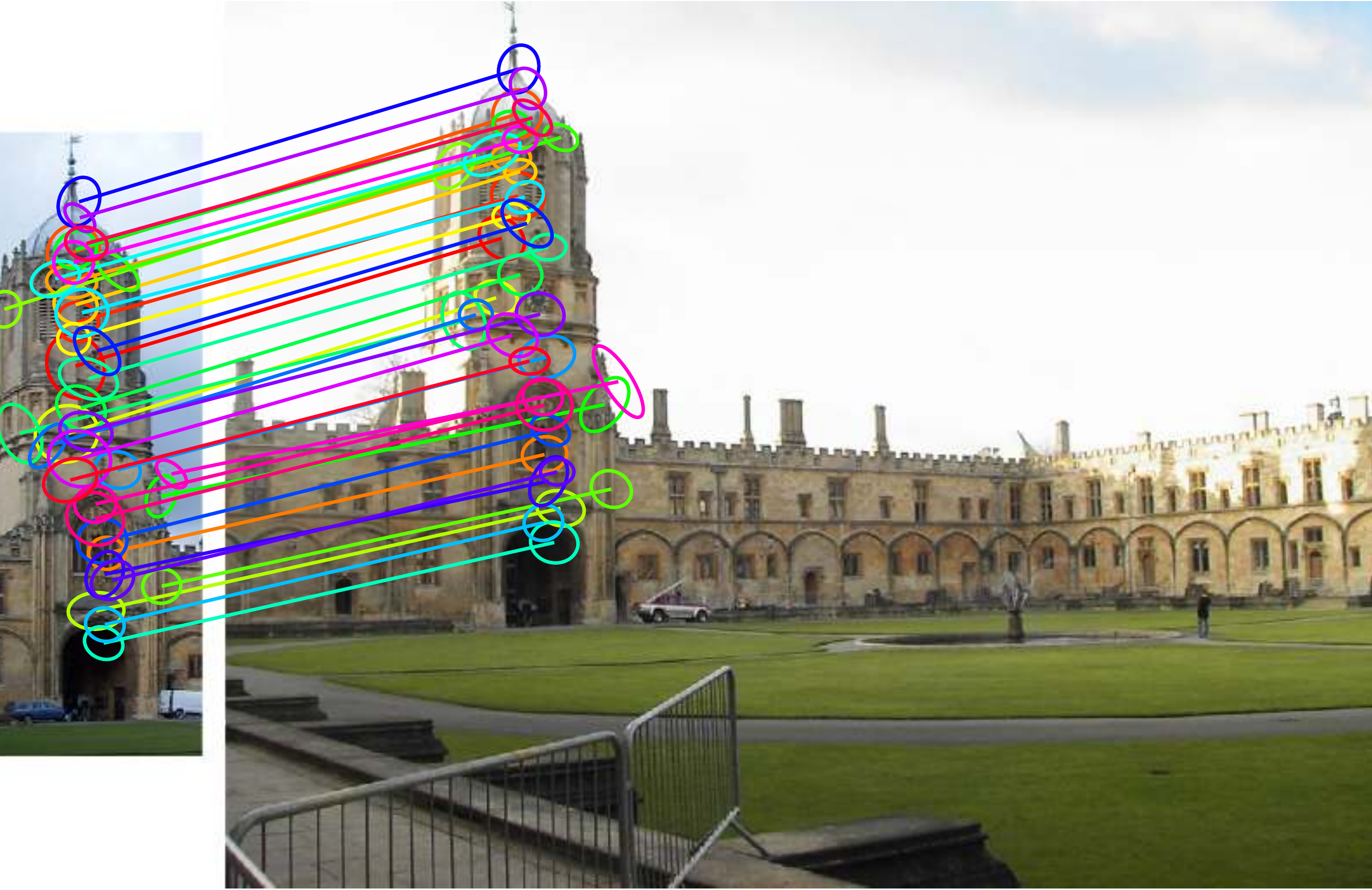} &
	\includegraphics[height=3.2cm]{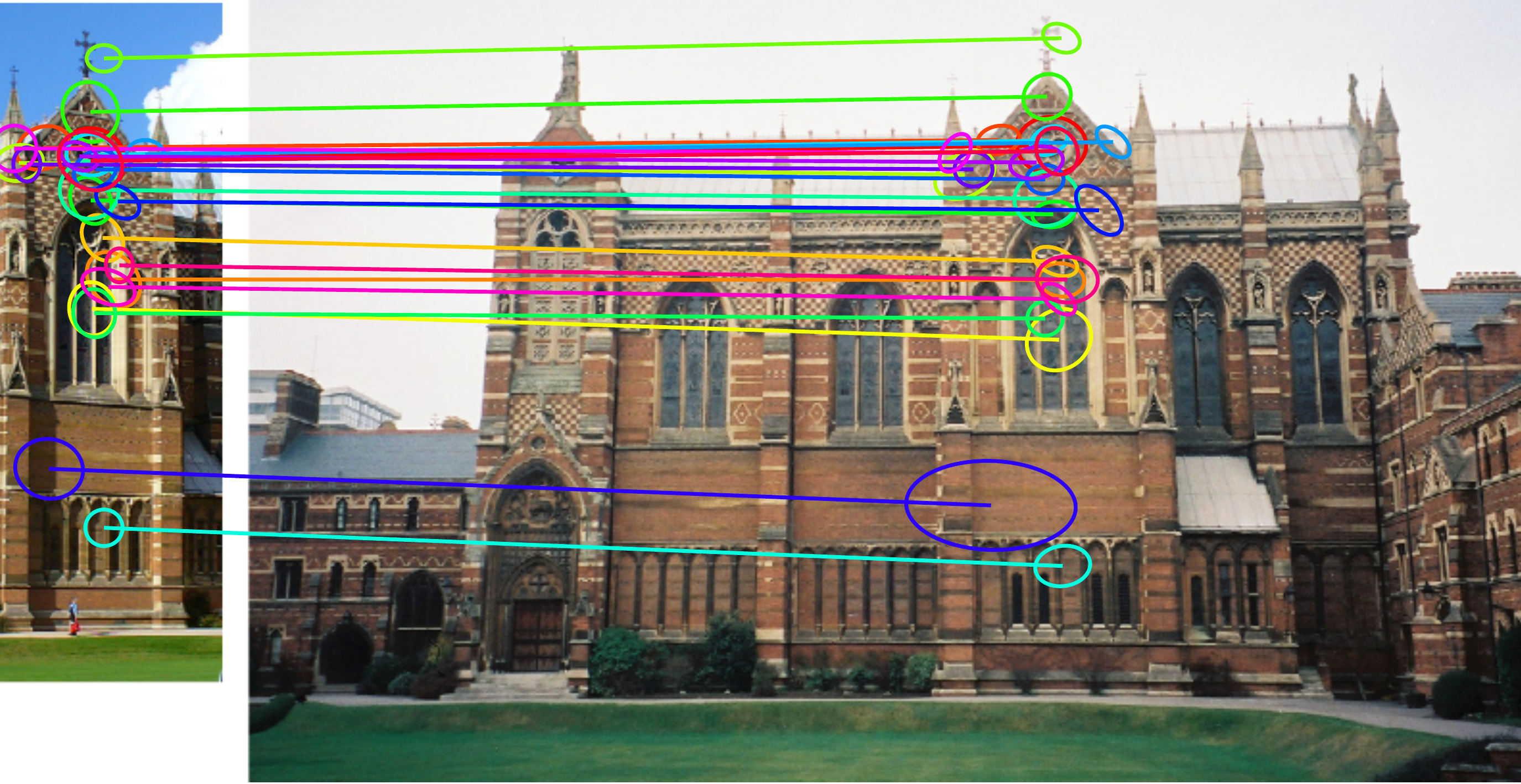}
\end{tabular}
\caption{Examples of our \emph{deep spatial matching} (DSM) between images from \roxf and \rpar benchmarks. Inlier features (ellipses) and correspondences (lines) shown in different colors.}\label{fig:soa}
\end{figure*}

\subsection{Local feature representation}
\label{sec:repr}

For each MSER $R$ detected in channel $j$ we compute a scalar value $v$ representing strength. It is pooled over the spatial support of $R$ in feature map $A^{(j)}$ as $v \defn \pool_{r \in R} A^{(j)}(r)$. Here $\pool$ can be any pooling operation like $\max$, mean, or generalized mean. We also fit an ellipse by matching its first and second moments, \ie its $2 \times 1$ mean (position) vector $\mu$ and $2 \times 2$ covariance matrix (local shape) $\Sigma$. For instance, Figure \ref{fig:fm_ell} shows an example of ellipses fitted to the MSER detected on feature maps of one channel for two views of the Oxford Museum of Natural History. Ellipses are well aligned in the two views. The local feature corresponding to $R$ is then represented by tuple $p \defn (\mu,\Sigma,v)$. Finally, we collect local features $\cP = (P^{(1)},\dots,P^{(k)})$ where $P^{(j)}$ contains the local features $p$ found in channel $j$. The entire operation is denoted by $\cP \defn d(A)$.

To treat feature channels as visual words, we assume
that features are uncorrelated, which does not hold in practice as indicated by the fact that whitening boosts performance. The same filter may respond to a variety of input patterns and worse, several filters may respond to the same pattern.
This can increase the level of interference in negative image pairs. For this reason we apply \emph{non-maximum suppression} (NMS) over all channels on the detected regions of each database image. Because local features are often small, we set a low IoU threshold. We do not apply NMS to the query image in order to allow matches from any channel.

\subsection{Spatial matching}
\label{sec:match}

Given the local features $\cP_1,\cP_2$ of two images $x_1,x_2$, we use \emph{fast spatial matching} (FSM)~\cite{PCISZ07} to find the geometric transformation $T$ between the two images and the subsets of $\cP_1,\cP_2$ that are consistent with this transformation. Matching is based on \emph{correspondences}, \ie pairs of local features $c = (p_1,p_2)$ from the two images. We allow pairs only between local features of the same channel, that is, $p_1,p_2$ are in $\cP_1^{(j)},\cP_2^{(j)}$ respectively for some channel $j$. We thus treat channels as \emph{visual words}, as if local features were assigned descriptors that were vector-quantized against a vocabulary and matched with the discrete metric. We begin with the \emph{tentative correspondences} that is the set of all such pairs, $\cC \defn (\cP_1^{(1)} \times \cP_2^{(1)}, \dots, \cP_1^{(k)} \times \cP_2^{(k)})$.

FSM is a variant of RANSAC~\cite{FB81} that generates a transformation hypothesis from a single correspondence. We adopt the linear 5-dof transformation which allows for translation, anisotropic scale and vertical shear but no rotation, assuming images are in ``upright'' orientation. Given a correspondence of two features $p_1=(\mu_1,\Sigma_1,v_1)$ and $p_2=(\mu_2,\Sigma_2,v_2)$, one finds from the two ellipses $(\mu_1,\Sigma_1), (\mu_2,\Sigma_2)$ the transformations $T_1,T_2$ that map them to the unit circle while maintaining the $y$-direction, and defines the transformation hypothesis $T = T_2^{-1} T_1$.

A hypothesis is evaluated based on the number of \emph{inliers}, that is, correspondences that are consistent with it. Because tentative correspondences are not too many, all possible hypotheses are enumerated. Following~\cite{PCISZ07}, we are using LO-RANSAC~\cite{ChMK03}, which iteratively evaluates promising hypotheses by fitting a full transformation to inliers by least squares. The transformation $T$ with the most inliers $\cM$ is returned. The operation is denoted by $(\cM,T) \defn g(\cP_1,\cP_2)$ and $\cM = (\cM^{(1)},\dots,\cM^{(k)})$ is a list of sets of inliers, one per channel.

\subsection{Retrieval and re-ranking}
\label{sec:rerank}

In an image retrieval scenario, $n$ database images $X = \{x_1,\dots,x_n\}$ are given in advance. For each image $x_i$ with feature tensor $A_i$, its local features $\cP_i \defn d(A_i)$ are computed along with a global descriptor $z_i$ spatially pooled directly from $A_i$ again \eg by max or \Gem pooling; $A_i$ is then discarded. At query time, given query image $x$ with feature tensor $A$, local features $\cP \defn d(A)$ and global descriptor $z$, we first rank $\{z_1,\dots,z_n\}$ by cosine similarity to $z$, and then the top-ranking images undergo spatial matching against $\cP$ according to $(\cM_i,T_i) \defn g(\cP,\cP_i)$
and are re-ranked according to \emph{similarity function} $s(\cM_i)$. The most common choice, which we also follow in this work, is the number of inliers found, $s(\cM_i) \defn \sum_{j=1}^k |\cM_i^{(j)}|$.

In order to improve the performance, we follow a \emph{multi-scale} approach where we compute feature tensors and local features from each input image at $3$ different scales, but still keeping a fixed number of local features from all scales according to strength. During re-ranking, we then perform spatial matching on all $9$ combinations of query and database image scales and keep the combination with maximum similarity.
Matching examples are shown in Figure \ref{fig:soa}.
As post-processing, we apply \emph{supervised whitening} to global descriptors as in~\cite{RTC18} and
query-time
\emph{diffusion}~\cite{ITA+16}. The latter is based on a nearest neighbor graph of the entire dataset $X$ and is a second re-ranking process applied after spatial re-ranking. The precision of top-ranking images is important for diffusion~\cite{RIT+18}, so spatial re-ranking is expected to help more
its presence.
\section{Experiments}
\label{sec:exp}

In this section we evaluate the benefits of different parts of our \emph{deep spatial matching} (DSM) and compare our results with the state of the art on standard benchmarks.

\subsection{Experimental setup}

\head{Test sets.} We use the medium and hard setups of the revisited \roxf and \rpar benchmarks~\cite{RIT+18}. We also use the large-scale benchmarks \roxfdist and \rpardist, which are a combination of a set of 1M distractor images with the two small ones. We resize all images to a maximum size of 1024$\times$1024. We evaluate performance by \emph{mean average precision} (\Map) and \emph{mean precision at 10} (\Mpr), as defined by the protocol~\cite{RIT+18}.

\head{Networks.}
We use VGG16~\cite{SZ14} and Resnet101~\cite{HZRS16}, denoted simply as \Vgg (\Res), or \V~(\R) for short. In particular we use the versions trained by Radenovic \etal~\cite{RTC18} with \Gem pooling. We also re-train them with max-pooling, on the same dataset of 120k Flickr images and the same structure-from-motion pipeline~\cite{RTC18}. Max-pooling is denoted by \Mac~\cite{TSJ15} and re-training by \re. \Res has a resolution 4 times smaller than \Vgg. Therefore we remove the stride in the first $conv5$ convolutional layer and add a dilation factor of 2 in all following layers. We thus preserve the feature space while upsampling by 2. This upsampling requires no re-training and is denoted by \up.

\head{Global image representation.}
To rank images based on cosine similarity, we compute the multi-scale global representation described in section~\ref{sec:match}. We extract descriptors at three different scales, related by factors $1$, $1/\sqrt{2}$, and $1/2$, and pooled from the last activation maps using max-pooling~\cite{TSJ15} (\Mac) or generalized mean-pooling~\cite{RTC18} (\Gem). The descriptors are pooled over scales into a single representation by either \Gem for networks using \Gem, or average for networks using \Mac.

\begin{table}
\begin{center}
\small
\setlength\tabcolsep{2pt}
\begin{tabular}{ |l|cc|cc|}
	\hline
	\multirow{2}{*}{Method}
		& \multicolumn{2}{c|}{Medium}
		& \multicolumn{2}{c|}{Hard} \\
	\cline{2-5}
		& \multicolumn{1}{c}\roxf
		& \multicolumn{1}{c|}\rpar
		& \multicolumn{1}{c}\roxf
		& \multicolumn{1}{c|}\rpar  \\
	\hline\hline
	\R-\mac\re
		& 64.0  
		& 75.5  
		& 36.7  
		& 53.2 \\  
	\R-\mac\re\up
		& 63.9   
		& 75.5   
		& 35.6   
		& 53.3  \\ 
	\R-\gem\cite{RTC18}
		& 64.7  
		& 77.2  
		& 38.5  
		& 56.3 \\  
	\R-\gem\cite{RTC18}\up
		& 65.3   
		& 77.3   
		& 39.6   
		& 56.6  \\ 
	\hline
	\hline
	\R-\mac\re+\D
		& 73.7  
		& 89.5  
		& 45.8  
		& 80.5  \\ 
	\R-\mac\re\up+\D
		& 73.9   
		& 89.9   
		& 45.6   
		& 81.0  \\ 
	\R-\gem\cite{RTC18}+D
		& 69.8 
		& 88.9  
		& 40.5  
		& 78.5  \\ 
	\R-\gem\cite{RTC18}\up+\D
		& 70.1   
		& 89.1   
		& 41.5   
		& 78.9   \\ 
	\hline
\end{tabular}

\caption{Impact of \Res (\R) activation upsampling (\up) on \Map in \roxf and \rpar~\cite{RIT+18}. \Mac: max-pooling~\cite{TSJ15}; \Gem: generalized-mean pooling~\cite{RTC18}; \D: diffusion~\cite{ITA+16}. All results with supervised whitening~\cite{MiMa07}. Citation specifies the origin of the network or \re: our re-training.}
\label{tab:resnet}
\end{center}
\end{table}

\begin{table}
\begin{center}
\small
\setlength\tabcolsep{1.8pt}
\begin{tabular}{ |l|cc|cc|cc|cc|cc|}
	\hline
		& \multicolumn{4}{c|}{Medium}
		& \multicolumn{4}{c|}{Hard} \\
	\cline{2-9}
		Method
		& \multicolumn{2}{c|}\roxf
		& \multicolumn{2}{c|}\rpar
		& \multicolumn{2}{c|}\roxf
		& \multicolumn{2}{c|}\rpar  \\
	\cline{2-9}
		& \map
		& \mpr
		& \map
		& \mpr
		& \map
		& \mpr
		& \map
		& \mpr \\
	\hline\hline
	\V
		& 44.8  
		& 63.3  
		& 65.7  
		& 95.0  
		& 18.4  
		& 31.2  
		& 41.0  
		& 79.1  \\ 
	\V+\dsm
		& 51.1  
		& 77.3  
		& 66.2  
		& 96.9  
		& 25.3  
		& 40.3  
		& 41.0  
		& 81.7  \\ 
	\R\up
		& 44.4  
		& 64.2  
		& 69.0  
		& 96.4  
		& 17.7  
		& 31.2  
		& 46.5  
		& 85.3  \\ 
	\R\up+\dsm
		& 49.6  
		& 74.0  
		& 69.7  
		& 98.4  
		& 21.7  
		& 37.6  
		& 46.7  
		& 87.0  \\ 
	\hline
	\hline
	\V+\D
		& 48.4  
		& 65.2  
		& 81.4  
		& 95.6  
		& 24.8  
		& 37.1  
		& 67.1  
		& 93.0  \\ 
	\V+\dsm+\D
		& 61.6  
		& 81.0  
		& 82.8  
		& 97.6  
		& 35.5  
		& 48.1  
		& 68.7  
		& 95.9  \\ 
	\R\up+\D
		& 53.8  
		& 69.0  
		& 85.6  
		& 96.3  
		& 29.8  
		& 38.1  
		& 72.1  
		& 94.1  \\ 
	\R\up+\dsm+\D
		& 60.2  
		& 78.9  
		& 86.3  
		& 96.9  
		& 33.1  
		& 42.0  
		& 72.8  
		& 95.0  \\ 
	\hline
\end{tabular}

\caption{Impact of the proposed \emph{deep spatial matching} (\Dsm) on \Map and \Mpr on \roxf and \rpar~\cite{RIT+18} with \emph{off-the shelf}
	(pre-trained on Imagenet~\cite{DSLLF09}) \Vgg (\V) and \Res (\R). \up: upsampling; \D: diffusion~\cite{ITA+16}. \Dsm: this work. All results with \alert{\Gem pooling} and supervised whitening.} 
\label{tab:off-shelf}
\end{center}
\end{table}

\head{Local feature detection.}
We use the MSER implementation of VLFEAT~\cite{VF8} to detect regions in the last activation map of the network. We set the minimum diversity to 0.7 and maximum variation to 0.5. We observed that the step $\Delta$ needs adjusting according to the network/dataset used. We do this by setting $\Delta$ to 60\% of the cumulative histogram of the activation values over the dataset.

\head{Local image representation.}
To spatially verify images, we compute the multi-scale local representation introduced in section~\ref{sec:match}. We fit an ellipse to each MSER region and for each ellipse we keep the covariance matrix, center position, channel id and maximum value. We discard activation maps with more than 20 features detected on query images, and 10 on database images. We apply NMS to features of database images with IoU threshold 0.2, which is restrictive enough even for small features. We rank features over all scales according to activation value and we select the top-ranking 512 features on \Vgg and 2048 on \Res.

\head{Re-ranking.}
After initial ranking by cosine similarity, we perform spatial matching between the query and the 100 top-ranked images as described in section~\ref{sec:match}. Tentative correspondences originate from the same channels. We set the error threshold to 2 pixels (in the activation channel, not the image) and the maximal scale change to 3. Finally, we use the number of inliers to re-rank the top 100 images.

\head{Spatially verified diffusion.}
We use diffusion~\cite{ITA+16}, denoted by \D, as a second post-processing step after spatial verification. It is based on a nearest neighbor graph of the global descriptors of the entire dataset, which is computed off-line. It starts from the top ranked images and finds more similar images according to manifold similarity. Diffusion is very powerful but sensitive to the quality of the initial top-ranked results. Thanks to our spatial matching, these results are more accurate. We take our 10 top-ranking spatially verified images and we compute a new score that is the product of the number of inliers and the descriptor similarity scores. We select the top 5 of them to initiate diffusion.

\subsection{Ablation experiments}

\head{Upsampling.}
Table~\ref{tab:resnet} shows the effect of upsampling on retrieval. This is not significant on \Mac pooling. On \Gem however, it results in perfomance increase by up to 1 \Map point on the hard setup of \roxf, both with and without diffusion. This can be explained by the higher resolution of the activation maps.

\head{Off-the-shelf networks.}
Our re-ranking can be applied to any network, even as pre-trained on Imagenet~\cite{DSLLF09} (\emph{off-the-shelf}). We use \Gem pooling, which is better than \Mac on such networks~\cite{RIT+18}. Table~\ref{tab:off-shelf} shows the effect of \Dsm on \roxf and \rpar medium and hard setup. We improve results with and without diffusion. The gain is significant on \roxf, up to 13 \Map points on \Vgg-\Gem with diffusion, medium setup. It is much smaller on \rpar, where the perfomance is already 20 to 40 \Map points higher than on \roxf.

\head{Whitening.}
We investigate the efficiency of our re-ranking with multi-scale global descriptors that are whitened or not. We use supervised whitening as in~\cite{MiMa07,RTC16}, denoted by \W. This is more powerful than PCA whitening~\cite{JC12}. As shown in Table~\ref{tab:whi}, we improve significantly on non-whitened descriptors with both networks on \roxf. We gain 3 to 4 \Map points, as well as increasing \Mpr. On the other hand, whitening boosts cosine similarity search, and gains 5 to 10 \Map points. Our improvement is more marginal \alert{or we lose up to one \Map point} in this case.

\begin{table}
\begin{center}
\small
\setlength\tabcolsep{1.5pt}
\begin{tabular}{ |l|cc|cc|cc|cc|cc|}
	\hline
		& \multicolumn{4}{c|}{Medium}
		& \multicolumn{4}{c|}{Hard} \\
	\cline{2-9}
		Method
		& \multicolumn{2}{c|}\roxf
		& \multicolumn{2}{c|}\rpar
		& \multicolumn{2}{c|}\roxf
		& \multicolumn{2}{c|}\rpar  \\
	\cline{2-9}
		& \map
		& \mpr
		& \map
		& \mpr
		& \map
		& \mpr
		& \map
		& \mpr \\
	\hline\hline
	\V\re
		& 55.2   
		& 78.1   
		& 61.3   
		& 96.1   
		& 25.0   
		& 38.6   
		& 35.8   
		& 77.4  \\ 
	\V\re+\dsm
		& 58.2   
		& 83.4   
		& 61.9   
		& 98.9   
		& 28.4   
		& 46.6   
		& 36.2   
		& 80.4  \\ 
	\V\re+\W
		& 59.1   
		& 81.3   
		& 66.8   
		& 97.7   
		& 31.5   
		& 49.0   
		& 41.7   
		& 82.3  \\ 
	\V\re+\W+\dsm
		& 60.0   
		& 84.3   
		& 67.0   
		& 98.6   
		& 32.5   
		& 53.1   
		& 42.0   
		& 82.3  \\ 
	\hline
	\hline
	\R\re\up
		& 54.0   
		& 75.7   
		& 70.6   
		& 97.0   
		& 24.2   
		& 36.6   
		& 44.4   
		& 84.6  \\ 
	\R\re\up+\dsm
		& 57.4   
		& 80.4   
		& 70.9   
		& 98.7   
		& 28.4   
		& 42.6   
		& 44.3   
		& 84.9  \\ 
	\R\re\up+\W
		& 63.9   
		& 85.2   
		& 75.5   
		& 98.4   
		& 35.6   
		& 52.6   
		& 53.3   
		& 89.6  \\ 
	\R\re\up+\W+\dsm
		& 62.7   
		& 83.7   
		& 75.7   
		& 98.7   
		& 35.4   
		& 51.6   
		& 53.1   
		& 88.6  \\ 
	\hline
\end{tabular}

\caption{
Impact of the \emph{supervised whitening} (\W)~\cite{MiMa07}
on \Map and \Mpr on \roxf and \rpar~\cite{RIT+18}.
Results with \Vgg (\V) and \Res (\R), both with \alert{\Mac pooling}; \up: upsampling; \D: diffusion~\cite{ITA+16}; \Dsm: this work; \re: our network re-training.}
\label{tab:whi}
\end{center}
\end{table}

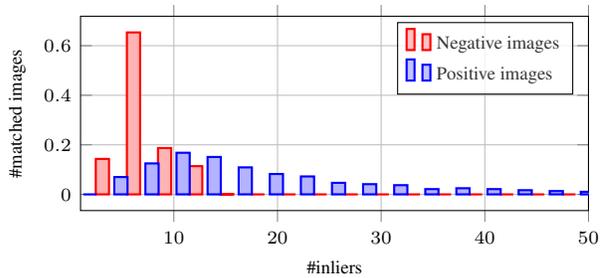
\begin{figure}
\centering
\extfig{inlier_dist}{\begin{tikzpicture}
\begin{axis}[%
	width=\columnwidth,
	height=0.5\linewidth,
	xlabel={\#inliers},
	ylabel={\#matched images},
	legend cell align={left},
	legend pos=north east,
    xmax = 50,
    xmin = 1,
    grid=both,
    line/.style={solid, mark=*,mark size=1.5,line width=1.0},
    ybar,
    bar width=5pt,
    ]

\pgfplotstableread{
bin	pos	neg
1	0	0.0041
4	0.07	0.1431
7 0.125	0.6536
10	0.168	0.187
13	0.151	0.114
16	0.109	0.0009
19	0.0823	0
22	0.0724	0
25	0.0466	0
28	0.0412	0
31	0.0372	0
34	0.0213	0
37	0.0248	0
40	0.0213	0
43	0.0169	0
46	0.0134	0
49	0.0104	0
52	0.0084	0
55	0.0074	0
58	0.0109	0
61	0.0074	0
64	0.0030	0
67	0.0050	0
70	0.0030	0
73	0.002	0
76	0.001	0
79	0.001	0
82	0.0005	0
85	0.001	0
88	0.0015	0
}{\inlEval}
	\addplot[red,fill=red!30] table[x=bin,y=neg] {\inlEval};\leg{Negative images};
	\addplot[blue,fill=blue!30] table[x=bin,y=pos] {\inlEval};\leg{Positive images};
\end{axis}
\end{tikzpicture}

%
}
\caption{Distribution of number of inliers for positive and negative database images over all queries of \roxf, using \Vgg-\Mac.}
\label{fig:inliers}
\end{figure}

\head{Inliers.}
To evaluate the quality of matching, we check how many inliers are found for positive and negative images. In particular, Fig.~\ref{fig:inliers} shows the distribution of the number of inliers to all queries of \roxf with \alert{\Vgg-\Mac} for both positive and negative images. The distribution is similar over different networks and datasets. Negative images can be easily discriminated by having few inliers, but this may result in loosing positive ones. Contrary to conventional spatial matching, we do not use local descriptors. This is positive in terms of memory, but comes necessarily with lower quality of matches. 
However, the top-ranking spatially verified images \emph{per query} are indeed accurate as indicated by \Mpr, which is enough to initiate a better diffusion.

\begin{table*}
\begin{center}
\small
\setlength\tabcolsep{1.4pt}
\begin{tabular}{ |l|cc|cc|cc|cc|cc|cc|cc|cc|cc|}
	\hline
		& \multicolumn{8}{c|}{Medium}
		& \multicolumn{8}{c|}{Hard} \\
	\cline{2-17}
		Method
		& \multicolumn{2}{c|}\roxf
		& \multicolumn{2}{c|}\roxfdist
		& \multicolumn{2}{c|}\rpar
		& \multicolumn{2}{c|}\rpardist
		& \multicolumn{2}{c|}\roxf
		& \multicolumn{2}{c|}\roxfdist
		& \multicolumn{2}{c|}\rpar
		& \multicolumn{2}{c|}\rpardist \\
	\cline{2-17}
		& \map
		& \mpr
		& \map
		& \mpr
		& \map
		& \mpr
		& \map
		& \mpr
		& \map
		& \mpr
		& \map
		& \mpr
		& \map
		& \mpr
		& \map
		& \mpr \\
	\hline
	\hline
	``DELF-ASMK*+SP''~\cite{RIT+18}
		& 67.8 
		& 87.9   
		& 53.8   
		& 81.1   
		& 76.9  
		& 99.3  
		& 57.3   
		& 98.3   
		& 43.1  
		& 62.4   
		& 31.2   
		& 50.7   
		& 55.4 
		& 93.4   
		& 26.4    
		& 75.7  \\
	\R-RMAC\cite{GARL17}~\cite{RIT+18}
		& 60.9  
		& 78.1    
		& 39.3   
		& 62.1   
		& 78.9  
		& 96.9  
		& 54.8   
		& 93.9   
		& 32.4  
		& 50.0   
		& 12.5   
		& 24.9   
		& 59.4 
		& 86.1   
		& 28.0   
		& 70.0 \\
	\hline
	\V-\mac\cite{RTC16}
		& 58.4   
		& 81.1   
		& 39.7   
		& 68.6   
		& 66.8   
		& 97.7   
		& 42.4   
		& 92.6   
		& 30.5   
		& 48.0   
		& 17.9   
		& 27.9   
		& 42.0   
		& 82.9   
		& 17.7   
		& 63.7 \\
	\V-\mac\re
		& 59.1  
		& 81.3  
		& 40.2   
		& 68.1   
		& 66.8   
		& 97.7   
		& 42.1   
		& 92.0   
		& 31.5   
		& 49.0   
		& 17.8   
		& 28.4   
		& 41.7   
		& 82.3   
		& 17.4   
		& 63.6 \\
	\V-\mac\re+\dsm
		& 60.0   
		& 84.3   
		& 42.2   
		& 71.0   
		& 67.0   
		& 98.6   
		& 42.5   
		& 94.7   
		& 32.5   
		& 53.1   
		& 19.4   
		& 31.6   
		& 42.0   
		& 82.3   
		& 17.7   
		& 66.0 \\
	\R-\mac\re\up
		& 63.9   
		& 85.2   
		& 43.2   
		& 69.6   
		& 75.5   
		& 98.4   
		& 50.1   
		& 95.3 
		& 35.6   
		& 52.6   
		& 17.7   
		& 31.4   
		& 53.3   
		& 89.6   
		& 22.4   
		& 71.6  \\
	\R-\mac\re\up+\dsm
		& 62.7   
		& 83.7   
		& 44.4   
		& 72.3   
		& 75.7   
		& 98.7   
		& 50.4   
		& 96.4   
		& 35.4   
		& 51.6   
		& 20.6   
		& 32.3   
		& 53.1   
		& 88.6   
		& 22.7   
		& 72.1 \\
	\hline
	\V-\gem\cite{RTC18}
		& 61.9   
		& 82.7   
		& 42.6   
		& 68.1   
		& 69.3   
		& 97.9   
		& 45.4   
		& 94.1   
		& 33.7   
		& 51.0   
		& 19.0   
		& 29.4   
		& 44.3   
		& 83.7   
		& 19.1   
		& 64.9 \\
	\V-\gem\cite{RTC18}+\dsm
		& 63.0   
		& 85.5  
		& 43.9   
		& 72.9   
		& 69.2   
		& 98.4   
		& 45.4   
		& 94.7   
		& 34.5   
		& 54.0   
		& 19.9   
		& 32.9   
		& 43.9   
		& 82.7   
		& 19.5   
		& 67.6 \\
	\R-\gem\cite{RTC18}
		& 64.7   
		& 84.7   
		& 45.2   
		& 71.7   
		& 77.2   
		& 98.1   
		& 52.3   
		& 95.3   
		& 38.5   
		& 53.0   
		& 19.9   
		& 34.9   
		& 56.3   
		& 89.1   
		& 24.7   
		& 73.3 \\
	\R-\gem\cite{RTC18}\up
		& 65.3   
		& 86.3   
		& 46.1   
		& 73.4   
		& 77.3   
		& 98.3   
		& 52.6   
		& 95.4   
		& 39.6   
		& 54.6   
		& 22.2   
		& 36.4   
		& 56.6   
		& 89.4   
		& 24.8   
		& 73.6 \\
	\R-\gem\cite{RTC18}\up+\dsm
		& 65.3   
		& 87.1   
		& 47.6   
		& 76.4   
		& 77.4   
		& 99.1   
		& 52.8   
		& 96.7   
		& 39.2   
		& 55.3   
		& 23.2   
		& 37.9   
		& 56.2   
		& 89.9   
		& 25.0   
		& 74.6 \\
	\hline
	\hline
	\multicolumn{17}{|c|}{Diffusion} \\
	\hline
	``DELF-HQE+SP''~\cite{RIT+18}
		& 73.4   
		& 88.2   
		& 60.6   
		& 79.7   
		& 84.0   
		& 98.3   
		& 65.2   
		& 96.1   
		& 50.3   
		& 67.2   
		& 37.9   
		& 56.1   
		& 69.3  
		& 93.7   
		& 35.8   
		& 69.1  \\
	``DELF-ASMK*+SP''$\to$\Ddelf~\cite{RIT+18}
		& 75.0   
		& 87.9   
		& 68.7   
		& 83.6   
		& \textbf{90.5}   
		& 98.0   
		& \textbf{86.6}   
		& 98.1   
		& 48.3   
		& 64.0   
		& 39.4   
		& 55.7   
		& \textbf{81.2}  
		& 95.6   
		& \textbf{74.2}   
		& 94.6  \\
	\hline
	\V-\mac\re+\D
		& 67.7  
		& 86.1 
		& 56.8   
		& 78.6  
		& 85.6  
		& 97.6  
		& 78.6   
		& 96.4   
		& 39.8   
		& 51.1   
		& 29.4   
		& 46.0   
		& 73.9 
		& 94.1 
		& 62.4   
		& 91.9 \\
	\V-\mac\re+\dsm+\D
		& 72.0 
		& 90.6 
		& 59.2   
		& 80.1   
		& 86.4  
		& 98.9  
		& 79.3   
		& 97.1   
		& 43.9   
		& 56.0  
		& 32.0   
		& 47.4   
		& 75.1  
		& 95.4 
		& 63.4   
		& 92.9 \\
	\R-\mac\re\up+\D
		& 73.9  
		& 87.9 
		& 61.3   
		& 80.6   
		& 89.9  
		& 96.1 
		& 83.0   
		& 95.1   
		& 45.6   
		& 62.2  
		& 31.9   
		& 48.4   
		& 81.0 
		& 94.3 
		& 68.6   
		& 91.9 \\
	\R-\mac\re\up+\dsm+\D
		& \textbf{76.9}  
		& 90.7 
		& 65.7   
		& 83.9   
		& 90.1  
		& 96.4 
		& 84.0   
		& 95.3   
		& \textbf{49.4}   
		& 64.7  
		& 35.7   
		& 51.3   
		& \textbf{81.2}   
		& 93.3  
		& 70.1   
		& 92.6  \\
	\hline
	\V-\gem\cite{RTC18}+\D
		& 69.6  
		& 84.7 
		& 60.4   
		& 79.4   
		& 85.6  
		& 97.1 
		& 80.7   
		& 97.1   
		& 41.1   
		& 51.1  
		& 33.1   
		& 49.6   
		& 73.9 
		& 93.7 
		& 65.3   
		& 93.1 \\
	\V-\gem\cite{RTC18}+\dsm+\D
		& 72.8  
		& 89.0 
		& 63.2
		& 83.7   
		& 85.7  
		& 96.1 
		& 80.1   
		& 95.7   
		& 45.4   
		& 57.1  
		& 35.4   
		& 53.7   
		& 74.2  
		& 93.3 
		& 65.2   
		& 91.9 \\
	\R-\gem\cite{RTC18}+\D
		& 69.8  
		& 84.0 
		& 61.5   
		& 77.1   
		& 88.9  
		& 96.9 
		& 84.9   
		& 95.9   
		& 40.5   
		& 54.4  
		& 33.1   
		& 48.2   
		& 78.5 
		& 94.6 
		& 71.6    
		& 93.7 \\
	\R-\gem\cite{RTC18}\up+\D
		& 70.1  
		& 84.3 
		& 67.5   
		& 79.0   
		& 89.1  
		& 97.3 
		& 85.0   
		& 96.6   
		& 41.5   
		& 54.4  
		& 39.6   
		& 53.0   
		& 78.9 
		& 95.1 
		& 72.0   
		& 94.1 \\
	\R-\gem\cite{RTC18}\up+\dsm+\D
		& 75.0  
		& 89.6 
		& \textbf{70.2}   
		& 84.5   
		& 89.3  
		& 97.1 
		& 84.8   
		& 95.3   
		& 46.2   
		& 60.6  
		& \textbf{41.9}   
		& 54.9   
		& 79.3 
		& 95.1 
		& 72.0   
		& 93.4 \\
  	\hline
\end{tabular}

\caption{
\Map and \Mpr \emph{state-of-the-art} on the full benchmark~\cite{RIT+18}. We use \Vgg (\V) and \Res (\R), with \Mac or \Gem pooling. \up: upsampling; \re: our re-training; \D: diffusion~\cite{ITA+16}. \Dsm: this work. Results citing~\cite{RIT+18} are as reported in that work and are combining DELF~\cite{noh2017largescale}, ASMK*~\cite{TAJ15} and HQE~\cite{ToJe14}. SP: spatial matching~\cite{PCISZ07}; \Ddelf: diffusion on the graph obtained by~\cite{GARL17}. The remaining citations specify where we took the trained network from.}
\label{tab:soa}
\end{center}
\end{table*}

\subsection{Comparison with the state-of-the-art}

We conduct an extensive comparison of our method with baselines and additional state-of-the-art methods. All methods are tested on \roxf, \roxfdist, \rpar and \rpardist. We collect all results in Table~\ref{tab:soa}.

Most baselines are improved by re-ranking, and all experiments on \roxf show consitent increase in performance. However, re-ranking is not perfect, as seen in Fig.~\ref{fig:inliers}. In few cases the performance drops after re-ranking by up to one \Map point on \rpar, in particular with the upsampled \Res-\Gem.
We attribute the loss to two factors. One is a limited ``vocabulary'', based only on 512 or 2048 activation maps. The other is the fact that activation maps are highly correlated. This is exploited by whitening of the global descriptors, but tends to create correlated features.

The performance is improved significantly when diffusion is initiated from top-ranked spatially verified images. Diffusion only needs few relevant images, and we are able to provide these images thanks to spatial matching. We improve on most datasets, networks and pooling options in this case. The gain is more pronounced on \roxf, and is up to 5 \Map or 6 \Mpr points.

Finally, the proposed method with spatially verified diffusion outperforms approaches based on deep local features in a number of cases. In particular, we compare with the best performing and expensive version of DELF~\cite{noh2017largescale} proposed and evaluated by~\cite{RIT+18}.
Apart from spatial verification by~\cite{PCISZ07} on the 100 top-ranking images, this version is using two independent representations. One is ASMK*~\cite{TAJ15}, based on 128-dimensional descriptors of 1000 DELF features per image, and used for initial ranking. Another is a global descriptor obtained by \Res-RMAC~\cite{GARL17}, and used for diffusion (\Ddelf) after spatial verification as in this work. By contrast, our global and local representations are obtained from the same activation tensor, and we do not use any local descriptors or their quantized versions.

\section{Discussion}
\label{sec:discussion}

Our experiments validate that the proposed representation for spatial verification achieves state-of-art performance across a number of different datasets, networks and pooling mechanisms. This representation arises naturally in the existing convolutional activations of off-the-shelf or fine-tuned networks, without any particular effort to detect local features or extract local descriptors on image patches. It does not require any network modification or retraining. It is a significant step towards bridging the gap between global descriptors, which are efficient for initial ranking using nearest neighbor search, and local representations, which are compatible with spatial verification.

Of course, the activation channels are not the most appropriate by construction to replace a visual vocabulary. This means that our representation, while being very compact, is not as powerful as storing \eg hundreds of local descriptors per image. Nonetheless, we still demonstrate
  that it is enough to provide high-quality top-ranking images to initiate diffusion, which then brings excellent results.

\head{Acknowledgments}
This work was supported by the GA\v{C}R grant 19-23165S and the OP VVV funded project CZ.02.1.01/0.0/0.0/16\_019/0000765 ``Research Center for
Informatics".

{\small
\bibliographystyle{ieee}
\bibliography{tex/main}

\begin{thebibliography}{10}\itemsep=-1pt

\bibitem{almazan2018re}
Jon Almazan, Bojana Gajic, Naila Murray, and Diane Larlus.
\newblock Re-id done right: towards good practices for person
  re-identification.
\newblock {\em arXiv preprint arXiv:1801.05339}, 2018.

\bibitem{BL15}
Artem Babenko and Victor Lempitsky.
\newblock Aggregating deep convolutional features for image retrieval.
\newblock In {\em ICCV}, 2015.

\bibitem{BSCL14}
Artem Babenko, Anton Slesarev, Alexandr Chigorin, and Victor Lempitsky.
\newblock Neural codes for image retrieval.
\newblock In {\em ECCV}, 2014.

\bibitem{choy2016universal}
Christopher~B Choy, JunYoung Gwak, Silvio Savarese, and Manmohan Chandraker.
\newblock Universal correspondence network.
\newblock In {\em NIPS}, pages 2414--2422, 2016.

\bibitem{ChMK03}
Ond{\v{r}}ej Chum, Jiri Matas, and Josef Kittler.
\newblock Locally optimized ransac.
\newblock In {\em DAGM Symposium on Pattern Recognition}, page 236. Springer
  Verlag, 2003.

\bibitem{CPSIZ07}
Ond{\v{r}}ej Chum, James Philbin, Josef Sivic, Michael Isard, and Andrew
  Zisserman.
\newblock Total recall: Automatic query expansion with a generative feature
  model for object retrieval.
\newblock In {\em ICCV}, October 2007.

\bibitem{DSLLF09}
Wei Dong, Richard Socher, Li Li-Jia, Kai Li, and Li Fei-Fei.
\newblock Imagenet: A large-scale hierarchical image database.
\newblock In {\em CVPR}, June 2009.

\bibitem{dosovitskiy2015flownet}
Alexey Dosovitskiy, Philipp Fischer, Eddy Ilg, Philip Hausser, Caner Hazirbas,
  Vladimir Golkov, Patrick Van Der~Smagt, Daniel Cremers, and Thomas Brox.
\newblock Flownet: Learning optical flow with convolutional networks.
\newblock In {\em CVPR}, pages 2758--2766, 2015.

\bibitem{FB81}
Martin~A. Fischler and Robert~C. Bolles.
\newblock Random sample consensus.
\newblock {\em Communications of ACM}, 6(24):381--395, 1981.

\bibitem{GARL17}
Albert Gordo, Jon Almazan, Jerome Revaud, and Diane Larlus.
\newblock End-to-end learning of deep visual representations for image
  retrieval.
\newblock {\em IJCV}, 124(2):237--254, Sep 2017.

\bibitem{HZRS16}
Kaiming He, Xiangyu Zhang, Shaoqing Ren, and Jian Sun.
\newblock Deep residual learning for image recognition.
\newblock In {\em CVPR}, 2016.

\bibitem{ITA+16}
Ahmet Iscen, Giorgos Tolias, Yannis Avrithis, Teddy Furon, and Ondrej Chum.
\newblock Efficient diffusion on region manifolds: Recovering small objects
  with compact cnn representations.
\newblock In {\em CVPR}, 2017.

\bibitem{JC12}
Herv\'e J\'egou and Ondrej Chum.
\newblock Negative evidences and co-occurrences in image retrieval: The benefit
  of {PCA} and whitening.
\newblock In {\em ECCV}, October 2012.

\bibitem{JDSP10}
Herv\'e J\'egou, Matthijs Douze, Cordelia Schmid, and P. P\'erez.
\newblock Aggregating local descriptors into a compact image representation.
\newblock In {\em CVPR}, June 2010.

\bibitem{KMO16}
Yannis Kalantidis, Clayton Mellina, and Simon Osindero.
\newblock Cross-dimensional weighting for aggregated deep convolutional
  features.
\newblock In Gang Hua and Herv{\'e} J{\'e}gou, editors, {\em Computer Vision --
  ECCV 2016 Workshops}, pages 685--701, Cham, 2016. Springer International
  Publishing.

\bibitem{kim2017fcss}
Seungryong Kim, Dongbo Min, Bumsub Ham, Sangryul Jeon, Stephen Lin, and
  Kwanghoon Sohn.
\newblock Fcss: Fully convolutional self-similarity for dense semantic
  correspondence.
\newblock In {\em CVPR}, 2017.

\bibitem{KrSH12}
Alex Krizhevsky, Ilya Sutskever, and Geoffrey~E. Hinton.
\newblock Imagenet classification with deep convolutional neural networks.
\newblock In {\em NIPS}. 2012.

\bibitem{LoZD14}
Jonathan~L Long, Ning Zhang, and Trevor Darrell.
\newblock Do convnets learn correspondence?
\newblock In {\em NIPS}. 2014.

\bibitem{L99}
David~G Lowe.
\newblock Object recognition from local scale-invariant features.
\newblock In {\em ICCV}, pages 1150--1157, 1999.

\bibitem{MCMP02}
Jiri Matas, Ondrej Chum, U. Martin, and T. Pajdla.
\newblock Robust wide baseline stereo from maximally stable extremal regions.
\newblock In {\em BMVC}, pages 384--393, September 2002.

\bibitem{MiMa07}
Krystian Mikolajczyk and Jiri Matas.
\newblock Improving descriptors for fast tree matching by optimal linear
  projection.
\newblock In {\em CVPR}, 2007.

\bibitem{MS04}
Krystian Mikolajczyk and Cordelia Schmid.
\newblock Scale and affine invariant interest point detectors.
\newblock {\em IJCV}, 60(1):63--86, 2004.

\bibitem{NS06}
David Nist\'er and Henrik Stew\'enius.
\newblock Scalable recognition with a vocabulary tree.
\newblock In {\em CVPR}, pages 2161--2168, June 2006.

\bibitem{noh2017largescale}
Hyeonwoo Noh, Andre Araujo, Jack Sim, Tobias Weyand, and Bohyung Han.
\newblock Largescale image retrieval with attentive deep local features.
\newblock In {\em ICCV}, 2017.

\bibitem{PCM09}
Michal Perdoch, Ondrej Chum, and Jiri Matas.
\newblock Efficient representation of local geometry for large scale object
  retrieval.
\newblock In {\em CVPR}, June 2009.

\bibitem{PCISZ07}
James Philbin, Ond{\v{r}}ej Chum, Michael Isard, Josef Sivic, and Andrew
  Zisserman.
\newblock Object retrieval with large vocabularies and fast spatial matching.
\newblock In {\em CVPR}, June 2007.

\bibitem{RIT+18}
Filip Radenovi{\'c}, Ahmet Iscen, Giorgos Tolias, Yannis Avrithis, and
  Ond{\v{r}}ej Chum.
\newblock Revisiting oxford and paris: Large-scale image retrieval
  benchmarking.
\newblock In {\em CVPR}, 2018.

\bibitem{RTC16}
Filip Radenovi{\'c}, Giorgos Tolias, and Ond{\v{r}}ej Chum.
\newblock {CNN} image retrieval learns from bow: Unsupervised fine-tuning with
  hard examples.
\newblock In {\em ECCV}, 2016.

\bibitem{RTC18}
Filip Radenovi{\'c}, Giorgos Tolias, and Ond{\v{r}}ej Chum.
\newblock Fine-tuning cnn image retrieval with no human annotation.
\newblock {\em IEEE Trans. PAMI}, 2018.

\bibitem{RSMC14}
Ali~Sharif Razavian, Josephine Sullivan, Atsuto Maki, and Stefan Carlsson.
\newblock Visual instance retrieval with deep convolutional networks.
\newblock {\em arXiv preprint arXiv:1412.6574}, 2014.

\bibitem{rocco2018end}
Ignacio Rocco, Relja Arandjelovic, and Josef Sivic.
\newblock End-to-end weakly-supervised semantic alignment.
\newblock In {\em CVPR}, 2018.

\bibitem{RCA+18}
Ignacio Rocco, Mircea Cimpoi, Relja Arandjelovi{\'c}, Akihiko Torii, Tomas
  Pajdla, and Josef Sivic.
\newblock {Neighbourhood Consensus Networks}.
\newblock In {\em NIPS}, Montr{\'e}al, Canada, December 2018.

\bibitem{SCDVPB17}
Ramprasaath~R. {Selvaraju}, Michael {Cogswell}, Abhishek {Das}, Ramakrishna
  {Vedantam}, Devi {Parikh}, and Dhruv {Batra}.
\newblock Grad-cam: Visual explanations from deep networks via gradient-based
  localization.
\newblock In {\em 2017 IEEE International Conference on Computer Vision
  (ICCV)}, pages 618--626, Oct 2017.

\bibitem{SZ14}
Karen Simonyan and Andrew Zisserman.
\newblock Very deep convolutional networks for large-scale image recognition.
\newblock {\em ICLR}, 2014.

\bibitem{SZ03}
Josef Sivic and Andrew Zisserman.
\newblock {Video Google}: {A} text retrieval approach to object matching in
  videos.
\newblock In {\em ICCV}, 2003.

\bibitem{ToAv11}
Giorgos Tolias and Yannis Avrithis.
\newblock Speeded-up, relaxed spatial matching.
\newblock In {\em ICCV}, 2011.

\bibitem{TAJ13}
Giorgios Tolias, Yannis Avrithis, and Herv{\'e} J{\'e}gou.
\newblock To aggregate or not to aggregate: Selective match kernels for image
  search.
\newblock In {\em ICCV}, December 2013.

\bibitem{TAJ15}
Giorgos Tolias, Yannis Avrithis, and Herv{\'e} J{\'e}gou.
\newblock Image search with selective match kernels: aggregation across single
  and multiple images.
\newblock {\em IJCV}, 2016.

\bibitem{ToJe14}
Giorgos Tolias and Herve Jegou.
\newblock Visual query expansion with or without geometry: Refining local
  descriptors by feature aggregation.
\newblock {\em Pattern Recognition}, 2014.

\bibitem{TSJ15}
Giorgos Tolias, Ronan Sicre, and Herv{\'e} J{\'e}gou.
\newblock Particular object retrieval with integral max-pooling of cnn
  activations.
\newblock {\em ICLR}, 2016.

\bibitem{VF8}
A. Vedaldi and B. Fulkerson.
\newblock {VLFeat}: An open and portable library of computer vision algorithms.
\newblock \url{http://www.vlfeat.org/}, 2008.

\bibitem{yi2016lift}
Kwang~Moo Yi, Eduard Trulls, Vincent Lepetit, and Pascal Fua.
\newblock Lift: Learned invariant feature transform.
\newblock In {\em ECCV}, 2016.

\bibitem{ZKL+16}
Bolei Zhou, Aditya Khosla, Agata Lapedriza, Aude Oliva, and Antonio Torralba.
\newblock Learning deep features for discriminative localization.
\newblock In {\em CVPR}, June 2016.

\end{thebibliography}
}

\end{document}